\documentclass{article}
\usepackage{arxiv}

\usepackage{booktabs} 
\usepackage[ruled]{algorithm2e} 
\renewcommand{\algorithmcfname}{ALGORITHM}
\SetAlFnt{\small}
\SetAlCapFnt{\small}
\SetAlCapNameFnt{\small}
\SetAlCapHSkip{0pt}
\IncMargin{-\parindent}

\usepackage{natbib}
\usepackage{soul}
\usepackage{url}
\usepackage[utf8]{inputenc}
\usepackage{amsmath}
\usepackage{booktabs}
\usepackage{layouts}

\usepackage{bibentry}
\usepackage{microtype}
\usepackage{multicol}

\usepackage{amsmath}

\newcount\Comments
\newcount\ResolvedComments
\usepackage[usenames,dvipsnames]{color}
\definecolor{darkgreen}{rgb}{0,0.5,0}
\definecolor{purple}{rgb}{0.5,0,0.5}
\newcommand{\kibitz}[2]{\ifnum\Comments=1\textcolor{#1}{#2}\fi}

\newcommand{\resolved}[1] {\ifnum\ResolvedComments=1\textcolor{blue}{[#1]}\fi}

\usepackage{dsfont}
\usepackage{titlesec}

\usepackage{amsmath}
\usepackage{bm}
\DeclareMathOperator{\GammaDist}{Gamma}
\DeclareMathOperator{\BetaDist}{Beta}
\DeclareMathOperator{\UniDist}{Uniform}
\DeclareMathOperator{\Ber}{Ber}
\newcommand{\NormalDist}{\mathcal{N}}

\newcommand{\matching}{\mathcal{M}}
\newcommand{\submissions}{\mathcal{U}}
\newcommand{\graders}{\mathcal{V}}
\newcommand{\components}{\mathcal{C}}
\usepackage{tikz}
\usetikzlibrary{bayesnet}

\title{Better Peer Grading through Bayesian Inference}
\author{
	Hedayat Zarkoob\\
	Department of Computer Science\\
	University of British Columbia\\
	Vancouver, Canada\\
	\texttt{hzarkoob@cs.ubc.ca} \\
	\And
	Greg d'Eon \\
	Department of Computer Science\\
	University of British Columbia\\
	Vancouver, Canada\\
	\texttt{gregdeon@cs.ubc.ca} \\
	\And
	Lena Podina\\
	Cheriton School of Computer Science\\
	University of Waterloo\\
	Waterloo, Canada\\
	\texttt{lpodina@uwaterloo.ca } \\
	\And
	Kevin Leyton-Brown \\
	Department of Computer Science\\
	University of British Columbia\\
	Vancouver, Canada\\
	\texttt{kevinlb@cs.ubc.ca} \\
}

\date{\vspace{-1em}}

\begin{document}

\maketitle


\begin{abstract}
Peer grading systems aggregate noisy reports from multiple students to approximate a ``true'' grade as closely as possible. Most current systems either take the mean or median of reported grades; others aim to estimate students’ grading accuracy under a probabilistic model. This paper extends the state of the art in the latter approach in three key ways: 
(1) recognizing that students can behave strategically (e.g., reporting grades close to the class average without doing the work); (2) appropriately handling censored data that arises from discrete-valued grading rubrics; and (3) using mixed integer programming to improve the interpretability of the grades assigned to students. We show how to make Bayesian inference practical in this model and evaluate our approach on both synthetic and real-world data obtained by using our implemented system in four large classes. These extensive experiments show that grade aggregation using our model accurately estimates true grades, students' likelihood of submitting uninformative grades, and the variation in their inherent grading error; we also characterize our models' robustness.


\end{abstract}


\section{Introduction}
\label{sec:intro}

Peer grading is a powerful pedagogical tool. 
It benefits students by giving them exposure to others' perspectives; helping them to internalize evaluation criteria by applying them critically to peer work~\cite{lu2012online}; and offering them feedback from equal-status learners~\cite{topping2009}.
Just as importantly, it gives instructors a way to make classes more scalable by shifting (some) grading workload away from course staff; effectively, this again benefits students, by giving them more opportunities for their work to be evaluated.

In order for peer grading systems to be both useful to instructors and acceptable to students, they must produce grades that are sufficiently similar to those that an instructor would have given. This is a challenging task because individual peer graders will be biased (consistently give generous or harsh grades); noisy (the same grader could grade an assignment differently on different days); and potentially strategic (some students will enter insincere peer grades unrelated to a submission's quality if they can get away with it). Addressing these interrelated challenges has been a topic of academic study in Computer Science for at least the last two decades. 


The first methods for aggregating peer grades---and many others introduced more recently---produce \textit{point estimates} of each assignment's grade and each grader's quality
%
\citep{walsh2014peerrank,chakraborty2018incentivizing, prajapati2020swagrader,dealfaro14, hamer05}. 
%
%
At their best, methods that produce point estimates 
maximize the likelihood of the data given a model, e.g., by 
assigning each grader a ``reliability'' parameter and iteratively updating these parameters to best describe the reported grades. (At worst, they do not even maximize likelihood. In this case they can produce grades and reliabilities that are inconsistent with each other, such as giving high weights to graders who are judged unreliable.)
Even when they do maximize likelihood, such point estimates can be overly confident. This can matter for model accuracy: e.g., the data might show that only one of several students is reliable without offering evidence about which is which, making it likely that the model will commit to the wrong explanation. It can also limit the way such models are used in real classes: e.g., an instructor may not want to trust student grades until the system is \emph{confident} that a peer grader is reliable.

These problems can be addressed by inferring distributions over grade and reliability estimates rather than point estimates.  
A seminal paper due to \citet{piech2013tuned} introduced the first such system, using graphical models to simultaneously determine distributions over both grades for student submission and accuracy assessments for each grader.
Their core ``PG1'' model assumes that  each assignment has a latent true grade and each peer grader has a latent bias and reliability; these parameters can be estimated from grading data through Bayesian inference, producing posterior distributions (and, hence, confidence intervals) on each assignment's grade and each student's grading abilities.
Piech et al.\ also introduced PG2 and PG3 models that respectively permit graders' biases to change over time and students' grading reliability to be correlated with their own assignments' grades.
Follow-up work by others further extended these models to allow for more complex reliability-grade correlations~\cite[the PG4 and PG5 models of][]{mi2015probabilistic} and to more explicitly account for differences between a single grader's reported grades~\cite[the PG6 and PG7 models of][]{wang2019improving}.

A key issue in peer grading is that students are asked to expend effort in grading each other's work, and that it is difficult to assess whether they did expend this effort. For example, students can subvert systems that assess grading quality by comparing individual grades to each other if they coordinate on all reporting the same grade. In response to this issue, an extensive line of work in the mechanism design literature focuses on incentivizing high quality reporting in peer grading and other crowdsourcing environments~\cite{prelec2004bayesian,jurca2009mechanisms,jurca2005enforcing,faltings2012eliciting,witkowski2012robust,witkowski2013dwelling,radanovic2013robust,radanovic2014incentives,riley2014minimum,kamble2015truth,kong2016putting,shnayder2016informed, liu2018surrogate,goel2019deep, gao2019incentivizing,zarkoob2020report}. 
Work in this area is mostly centered around the idea of \emph{peer prediction}, developing mechanisms that incentivize graders to grade the peer carefully by rewarding them based on comparing their peer grades to others'.  
While our model of low-effort grading is inspired by work in this area, these mechanisms typically rely on restrictive modeling assumptions, making them inappropriate for most practical peer grading systems. 
Further, \citet{burrell2021measurement} found that rewards from out-of-the-box peer prediction mechanisms do not accurately reflect grader effort levels on realistic simulated data.

Despite the considerable intellectual progress just described, there remain obstacles to deploying AI-based peer grading systems in practice. 
Statistically rich methods based on graphical models and economically informed mechanism design approaches have been developed independently; we are not aware of any system that unifies the two by providing both Bayesian parameter estimates and meaningful incentives for students to invest effort in peer grading. Furthermore, the statistical literature allows both students and the peer grading system itself to assign real-valued grades, whereas real instructors tend to use coarse-valued rubrics, particularly when eliciting grades from students. This can harm inference and also requires the instructor to find a way of mapping real-valued grades back onto their course's rubric. Even if this mapping produces accurate grades, students must be able to understand it in order to trust the system~\cite{kizilcec2016trust}.

This paper addresses all of these problems, introducing extensions to probabilistic peer grading systems that can detect (and hence enable the disincentivization of) low-effort, strategic behavior by students; improve inference quality in the presence of discrete grading rubrics; and output interpretable, discrete final grades that closely approximate maximum a posteriori estimates. In what follows we begin by introducing notation and formally defining the baseline PG1 model (Section 2). We then introduce our novel methods for modeling grader effort, modeling discrete grade reports as censored observations, and outputting interpretable discrete grades via mixed-integer programming (Section 3). We evaluate our contributions in two ways. First, using real data from four offerings of a large class, we show that each of our effort and censoring model extensions improve likelihood on held-out data and that our method for generating interpretable grades closely tracks MAP estimates (Section 4). Second, using simulated data generated using the hyperparameters fit in the previous section, we assess our model's ability to recover grades and grader reliabilities as a function of dataset size and in the presence of hyperparameter misspecification (Section 5). We conclude by discussing ways in which our methods can be leveraged in the classroom (Section 6).\footnote{Open-source implementations of our models are available at \url{https://github.com/hezar1000/mta-inference-public}.}

\section{Technical Setup}
\label{sec:existing_models}
We will use the following notation throughout the paper. 
Let $\submissions$ be the set of submissions and $\graders$ be the set of graders.
It is common for instructors to provide graders with a \emph{rubric}: a decomposition of the overall grade into a set of separate components. 
To capture this, we assume that each submission is graded on $C$ components $\components = \{1, 2, \dots, C\}$, with a maximum grade of $M$ for each.
Lastly, we write $\NormalDist(\mu, \sigma^2)$ to denote the normal distribution with mean $\mu$ and variance $\sigma^2$.

\begin{figure}
    \centering
    \includegraphics[width=0.65\linewidth,trim={0 0.5cm 0 0}]{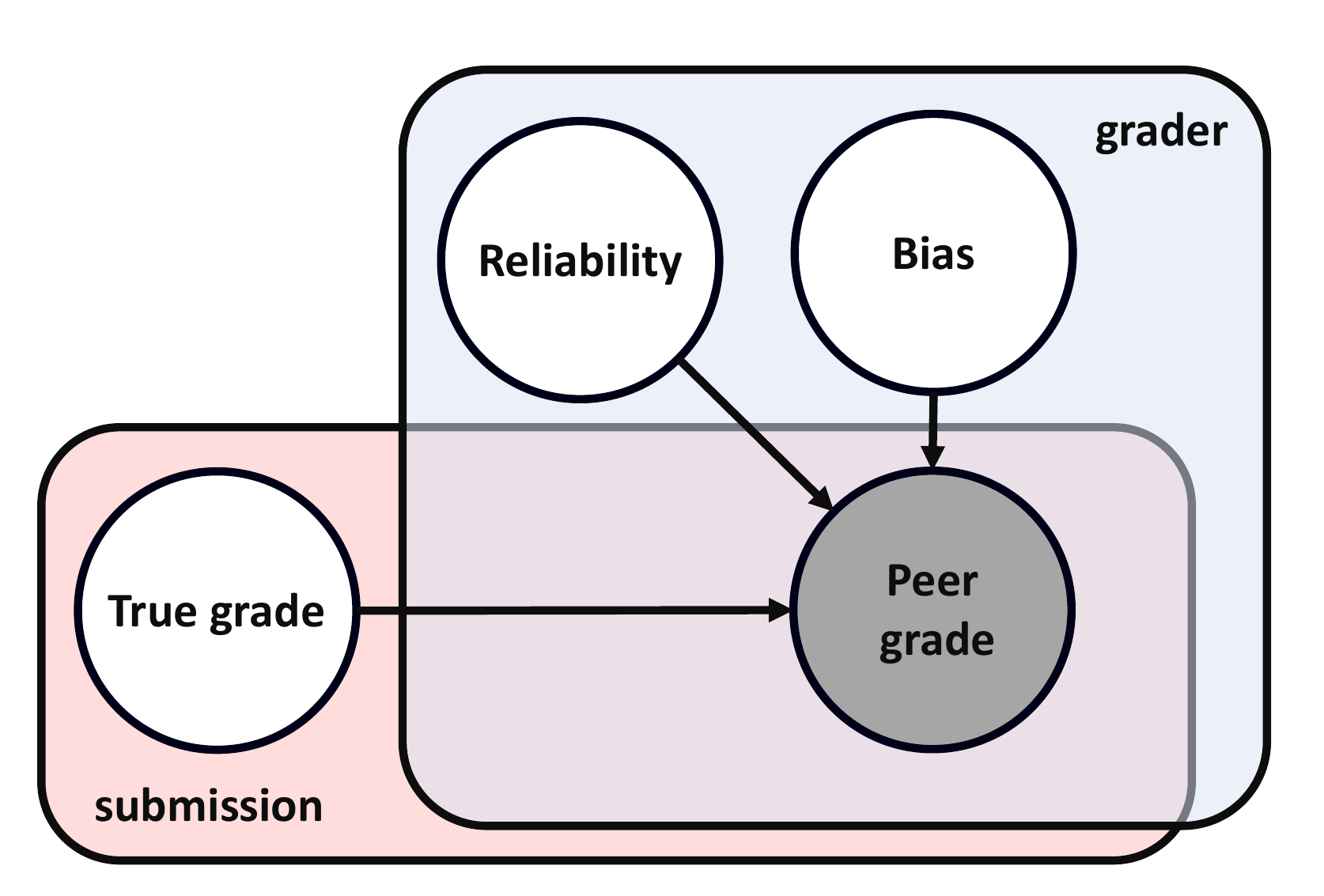}
    \caption{The PG1 graphical model.}
    \label{fig:PG1_graphical_model}
\end{figure}

We now define PG1~\cite{piech2013tuned}, a key graphical model from the literature, which models students as having reliabilities and biases.
The PG1 model supposes three sets of latent variables.
Each submission $u \in \submissions$ has a \emph{true grade} $s_{u,c} \in \mathds{R}$ for each rubric component $c \in \components$.
Each grader $v \in \graders$ is described by a \emph{reliability} $\tau^v \in \mathds{R}^+$, which captures the consistency of their grading, and a \emph{bias} $b^v \in \mathds{R}$, which describes their tendency to give generous or harsh grades.
An ideal grader would have a high reliability and 0 bias.
Then, when a grader $v$ grades a submission $u$, they give a peer grade $g_{u,c}^v \in \mathds{R}$ for each component $c$.
Concretely, the PG1 data generating process, depicted in Figure~\ref{fig:PG1_graphical_model}, is:
\begin{alignat*}{2}
    \text{(True grades)}& \quad 
    s_{u,c} &&\sim \mathcal{N}(\mu_s, 1 / \tau_s); \\
    \text{(Reliabilities)}& \quad 
    \tau^v &&\sim \GammaDist(\alpha_\tau, \beta_\tau); \\
    \text{(Biases)}& \quad 
    b^v &&\sim \mathcal{N}(0, 1 / \tau_b); \\
    \text{(Peer grades)}& \quad
    g_{u,c}^v &&\sim \mathcal{N}(s_{u,c} + b^v, 1 / \tau^v).
\end{alignat*}
This model has five hyperparameters: $\mu_s$ and $\tau_s$ fix the prior distribution of true grades; $\alpha_\tau$ and $\beta_\tau$ fix the prior over reliabilities (gamma-distributed to ensure that reliabilities are positive); and $\tau_b$ is the precision of the bias distribution.

Armed with a dataset of peer grades, the goal of the model is to infer true grades for each submission and reliabilities and biases for each grader.
Such a complex model does not give rise to a closed-form expression for the posterior distribution over these parameters.
Instead, the posterior must be estimated numerically. 
A good option is \emph{Gibbs sampling}~\cite{Geman1984Gibbs}: initializing each variable randomly, repeatedly sampling new beliefs about a single variable in the model conditional on beliefs about all other variables, and reporting the long-run distributions of these samples. 
This approach is particularly attractive for PG1, as the true grade, bias, and reliability priors are conjugate priors for the normally-distributed peer grade likelihoods, giving each of the Gibbs updates a simple closed form.
We present these update equations in Appendix~\ref{sec: gibbs_updates}.




\section{Methods}
\label{sec: inference}
We now present our main conceptual contributions.
First, we show how low-effort grading behavior can be disincentivized by augmenting the probabilistic model to include latent variables describing graders' \emph{effort}. 
Second, we show how to better handle the common case where graders select discrete grades from a coarse rubric by modeling these reports as \emph{censored} observations of an underlying real value. 
Third, we introduce a mixed-integer programming method for identifying \emph{interpretable} weighted averages of the peer grades that are faithful to the model's posterior beliefs.
We present and evaluate these features as extensions to PG1, which we found most applicable to our own class, but they could be applied to any of the PG* models in the literature. In Appendix~\ref{sec: PG5}, we show how these extensions could be added to the PG5 model of \citet{mi2015probabilistic}, additionally modeling correlations between students' submission grades and reliabilities.


\subsection{Modeling Grader Effort}
In order for a statistical model to be able to accurately recover true grades from peer reports, students must invest the effort required to grade as well as they can. Typically, students are incentivized to do so in part by receiving explicit grades for their peer grading prowess.  However, it is not easy to determine whether a student has done a good job of peer grading when there are no instructor or TA grades to which their evaluation can be compared. 
The main alternative method of providing an incentive---called peer prediction---is based on comparing students to each other. When all other students grade as accurately as they can, it is often possible to design reward systems that incentivize a given student to do the same (i.e., making effortful reporting an equilibrium). Unfortunately, however, other equilibria also exist in which students coordinate on the same grade without reading the assignment~\cite{jurca2009mechanisms, waggoner2014output, gao2014trick}. 

We can reduce students' incentives for such low-effort behavior by explicitly modeling it, helping us to avoid assigning high reliabilities to low-effort students. 
Each time a grader $v$ grades a submission $u$, we assume they make a binary decision about whether to make an \emph{effort} $z_u^v$ on the submission.
If they choose to make an effort, they produce a noisy grade as usual; otherwise, they choose a random grade from a fixed ``low-effort'' distribution $D_\ell$.
For simplicity, we model these effort decisions as being independent of the content of the submission.
Then, each student has an effort probability $e^v$, modeling their likelihood of exerting high effort when grading a submission.
Formally, adding this feature to PG1 produces the model: 
\begin{alignat*}{2}
    \text{(True grades)}& \quad 
    s_{u,c} &&\sim \mathcal{N}(\mu_s, 1 / \tau_s); \\
    \text{(Reliabilities)}& \quad 
    \tau^v &&\sim \GammaDist(\alpha_\tau, \beta_\tau); \\
    \text{(Biases)}& \quad 
    b^v &&\sim \mathcal{N}(0, 1 / \tau_b); \\
    \text{(Effort prob.)}& \quad 
    e^v &&\sim \BetaDist(\alpha_e, \beta_e); \\
    \text{(Efforts)}& \quad
    z_u^v &&\sim \Ber(e^v); \\
    \text{(Peer grades)}& \quad
    g_{u,c}^v &&\sim \begin{cases}
        \NormalDist(s_{u,c} + b^v, 1 / \tau^v), & z_u^v = 1; \\
        D_\ell, & z_u^v = 0.
    \end{cases}
\end{alignat*}
Regardless of our choice of $D_\ell$, Gibbs sampling remains straightforward: both efforts and effort probabilities yield closed-form updates, and all other parameter updates simply exclude grades that a given sample calls low effort.

So, which low-effort distribution $D_\ell$ should we choose?
\citet{hartline2020} showed that the most robust ``low effort'' strategy is to report the class average, minimizing the expected distance to an effortful report. 
A point-mass low-effort model would be extremely brittle, so a natural $D_\ell$ is a normal distribution centered on the class average. 
We would also prefer to model idiosyncratic grading strategies as low effort, such as assigning everything a low or high grade, so that these outlying grades do not drive our reliability estimates.
Thus, our final $D_\ell$ is a mixture between this normal distribution and a uniform distribution: 
\begin{alignat*}{2}
    D_\ell = \begin{cases}
        \NormalDist(\mu_s, 1 / \tau_\ell), & \text{with probability } 1-\epsilon; \\
        \UniDist(0, M), & \text{with probability } \epsilon.
    \end{cases}
\end{alignat*}
Adding effort to a model introduces four new, tunable hyperparameters: $\alpha_e$ and $\beta_e$ parameterize a prior over graders' effort probabilities, which is beta-distributed to ensure that these probabilities are between $0$ and $1$; $\tau_\ell$ describes the amount of noise in graders' reports when they put in low effort; and $\epsilon$ specifies the probability with which a low-effort grader reports a grade uniformly at random. 

\subsection{Discrete Rubrics as Censored Observations}
In practice, submissions are usually graded on coarse, discrete rubrics such as five-point scales (and virtually no class allows graders arbitrary decimal precision).
PG1 and all of its successors nevertheless assume that reported grades are real valued. They do this for two good reasons. First, continuous distributions like Gaussians are realistic models of true grade distributions, and arguably reported grades are just discretizations of the same continuous values. 
Second, most discretizations produce non-conjugate priors, making Gibbs updates computationally intractable.

However, failing to model grades as discrete-valued can skew a model's posterior beliefs. 
Treating discrete grades as real-valued can add statistical bias to graders' reliability estimates, both by overestimating (e.g., graders appear to be in perfect agreement when their rounded grades are interpreted as draws from a continuous distribution) or underestimating (e.g., graders can appear to disagree substantially after rounding, even if they correctly assess that a true grade is close to the midpoint between two integers). Of course, degraded reliability estimates lead to degraded true grade estimates.

\begin{figure}
    \centering
    \includegraphics[width=0.75\linewidth,trim={0 0.5cm 0 0}]{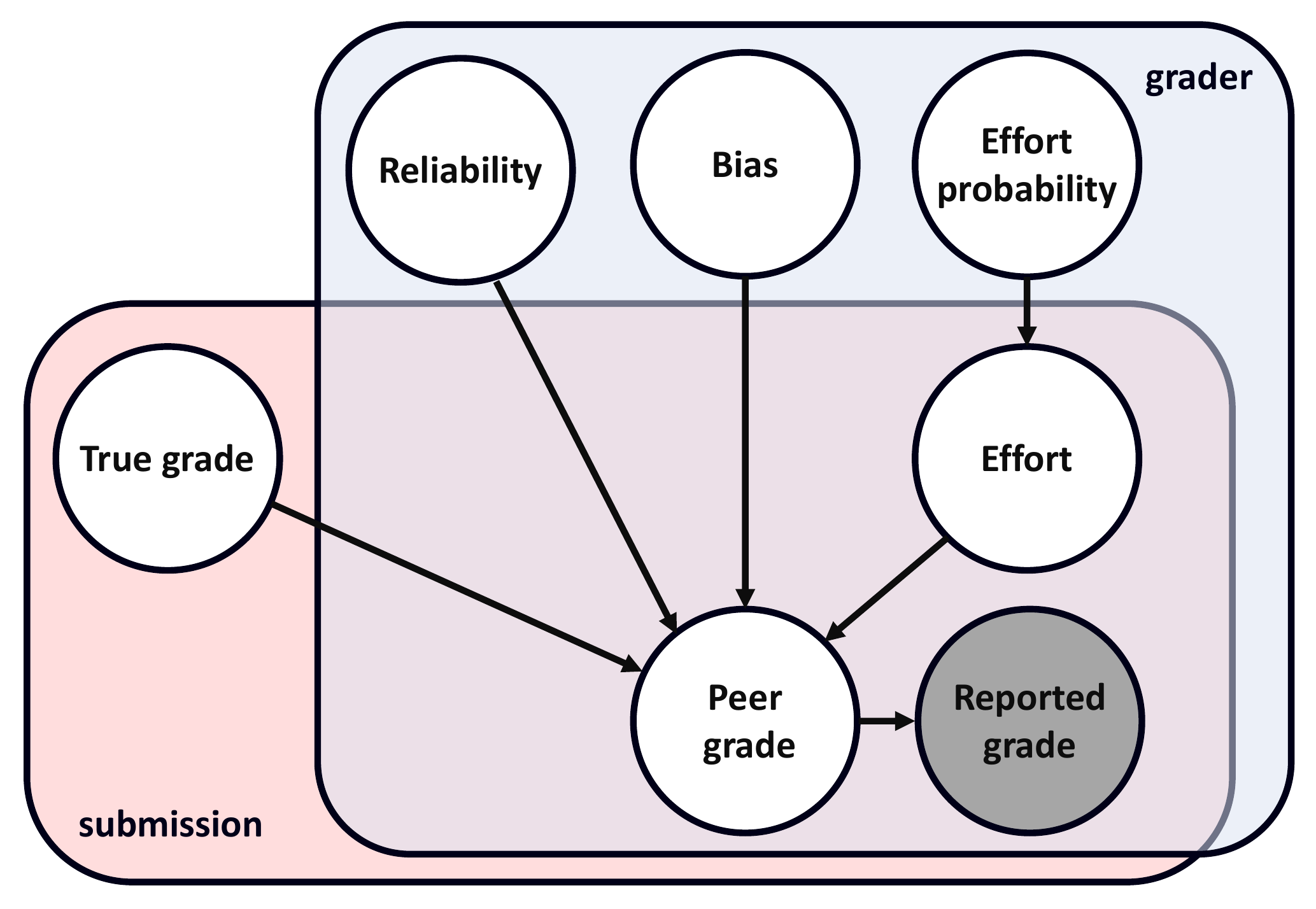}
    \caption{Our complete graphical model.}
    \label{fig:graphical_model}
\end{figure}

We propose an approach for extending PG-style models to realistic discrete grade distributions that maintains the tractability of Gibbs updates.
Let $G \subset \mathds{Z}$ be a set of legal discrete grades (e.g., integers between $0$ and $5$), and let $n_G: \mathds{R} \to G$ be a function mapping a grade to its nearest value in $G$, rounding up.
We continue to model a grader $v$ grading a submission $u$ as sampling a real-valued peer grade $g^v_{u,c}$, 
but we now treat these real-valued peer grades as latent variables, with the student reporting the discrete grade $r_{u,c}^v = n_G(g^v_{u,c})$: 
a \emph{censored observation} of the real-valued peer grade.
Figure~\ref{fig:graphical_model} shows the resulting graphical model. 
Observe that it correctly leads us to consider grader disagreement to be more likely when an assignment's true grade is 3.51 than when it is 3.0.

Naive Gibbs sampling would be extremely inefficient for this model: the posterior distribution has multiple modes where the true grade nearly matches one latent peer grade, and Gibbs sampling rarely moves between these modes.
To avoid this problem, we instead marginalize over peer grades, integrating over all of their possible values.
Although the Gibbs updates no longer have closed-form expressions, there is an efficient discrete approximation.
First, notice that we can still straightforwardly compute the likelihood of a reported grade for any setting of the submission's true grade and the grader's reliability and bias.
Then, to update a true grade variable, we consider a uniform grid of possible grades (ranging from 0 to a grade slightly above the maximum grade $M$), compute an unnormalized posterior probability for each value, renormalize these probabilities to sum to 1, and sample from the resulting discrete distribution.
The reliability and bias updates are similar, testing uniform grids of plausible reliabilities and biases.
We provide full details of these Gibbs updates in Appendix~\ref{sec: gibbs_updates}.

\subsection{Explaining Discrete Grades via MIP}

A key advantage of a Bayesian approach to reasoning about peer grades is that it yields distributional posterior beliefs about each quantity of interest rather than point estimates. However, students still expect to receive discrete grades rather than probability distributions. Furthermore, if the course staff grade on the same rubric as the students, assigning real-valued grades to students incentivizes half of them to ask for regrades (e.g., if their true grade is 3.6, the rounded TA grade would be 4). How should we turn posterior distributions into discrete grades? 
One might map the mean of the Gibbs samples to the nearest rubric element, but this can lead to rounding errors. A better option is to choose the rubric element corresponding to the continuous grade interval having the highest mass in the posterior distribution---the \emph{maximum a posteriori (MAP)} grade. While this approach is statistically sensible, it leaves students with little insight about how their peer grades influenced the calculation; this can lead to reduced trust in the system~\cite{kizilcec2016trust} and more appeals. 

Both approaches have an additional problem: they sometimes produce final grades larger or smaller than any peer grade (e.g., when the model assigns biases having the same sign to all graders). In our experience, students find such grades confusing and unfair; they instead expect to receive final grades that interpolate their received grades, such as averages weighted by each grader's perceived trustworthiness.
Students also tend to fixate on low-quality peer reviews, strongly preferring such reviews to have no impact on their final grade rather than a small impact.

We propose a novel mixed integer programming (MIP) formulation that maps posterior grade distributions to discrete final grades that can be explained as rounded weighted averages of reported peer grades.
To start, we assign a weight for each grader in proportion to their reliability and effort estimates.
The MIP then adjusts these weights in two ways to maximize the posterior probability of the resulting rounded weighted average.
First, we allow the MIP to deviate from each grader's initial weight by up to a constant $S$ to improve the likelihood of the resulting rounded grade. 
Second, we require the MIP to assign weights either equal to zero or above a minimum threshold $T$, avoiding small, non-zero weights on relatively uninformative grades.
Notice that the resulting weighted averages can never produce a grade outside of the range of the peer grades.
We define our MIP formulation formally in the Appendix~\ref{sec: mip}.

\section{Validation Experiments on Classroom Data}
\label{sec: real_data}
We now evaluate our contributions on real peer grading data, gathered between September 2018 and December 2021 from four offerings of an undergraduate-level computer science course on the ethical and societal impacts of computing. (Research use of this data was authorized by an ethics review board.)
In each offering of this course, approximately 120 students 
wrote 11 weekly essays.
Each grade consisted of discrete values between 0 and 5 for each of four components (structure, evidence, subject matter and English). 
Overall, our experiments show that our model extensions (grader effort and censored observations) improved fit and that the explainable grades output by our MIP rarely differed from the (non-explainable) MAP estimates.


Our experiments include data from three types of graders.
First, each student essay was graded by 4--5 peers, yielding between 6088 and 7068 peer grades per dataset. 
Second, each course was supported by a team of 3--5 TAs who spot checked between 474 and 644 essays, mostly in response to suspiciously high average grades, high disagreement between peer grades, and graders with poor historical performance. 
Our TAs were diligent and responsible, so we clamped their efforts to 1 (i.e., their grades could never be explained away as coming from the low-effort distribution); we fit their bias and reliability parameters from data just as we did for students.
Third, each course included between 60 and 84 gold-standard ``calibration'' submissions that we used to train students; their grades were painstakingly agreed upon by the whole course staff in special grading sessions.
We model these grades as having being given by a special ``instructor'' grader, with efforts clamped to 1 and a reliability clamped to 16 (corresponding roughly to an 80\% accuracy of perfectly recovering true grades); we fit its bias parameter from data.

Ideally, we would evaluate our models based on their ability to recover assignments' true grades.
However, we lacked enough ground-truth values for a sound evaluation: while TAs and instructors are reliable sources of grades, they graded an unrepresentative sample of the submissions, and we did not have the resources to obtain TA grades more broadly after the classes were finished.
Instead, we tested our models' ability to predict held out data---their \emph{held-out likelihood}~\cite{Vehtari2016}.
Running leave-one-out cross-validation would have been computationally prohibitive, given the size of our datasets and the cost of our inference procedure, so we instead used 10-fold stratified cross-validation.
We first split the dataset into 10 groups of $n/10$ peer grades, ensuring that no two peer grades on the same submission were in the same group.
Then, for each way of selecting 9 groups from the 10, we ran the model on these selected observations, summing the model's log likelihoods on the remaining group.
An exploratory experiment on one of our four datasets confirmed that this approach closely approximated leave-one-out cross-validation.
We use paired $t$-tests to make statistical comparisons between held-out likelihoods of several models on the same dataset.

Each time we fit our model to a dataset, we collected 4 runs of 1,100 Gibbs samples, discarding the first 100 burn-in samples from each run and concatenating the remainder; this took about 8 CPU hours.
We found that this number of samples made a good tradeoff between the runtime and sample complexity of our models: e.g., comparing to runs with 10,000 Gibbs samples, our protocol of gathering 4,000 samples caused an average error of 1\%, 3\%, and 1\%  on our estimates of true grades, reliabilities, and effort, respectively, and therefore had a small impact on our results.

In order to evaluate the effectiveness of our effort and censoring extensions to PG1, we compared models having both, one, or neither of these features.
We independently optimized each model's hyperparameters using randomized grid search, choosing the hyperparameters that maximized the model's held-out likelihood; full details of this hyperparameter search are presented in Appendix~\ref{sec: hyperparam_search}. 
We found that the model using both features had the highest held-out likelihood.
Furthermore, many of this model's hyperparameters had the same optimal values on all four classes, sharing the same true grade, 
reliability, 
and effort probability priors. 
They differed only (i)~in their bias priors, with two datasets giving rise to small biases ($\sigma_b = 0.1$) and two to larger biases ($\sigma_b = 1$); and (ii)~in their low-effort grading distributions, with low-effort reports being relatively diffuse in three datasets ($\tau_\ell = 1$) but more concentrated in the fourth ($\tau_\ell = 4$). 

\begin{figure}
    \centering
    \includegraphics[trim={0 0.4cm 0 0}]{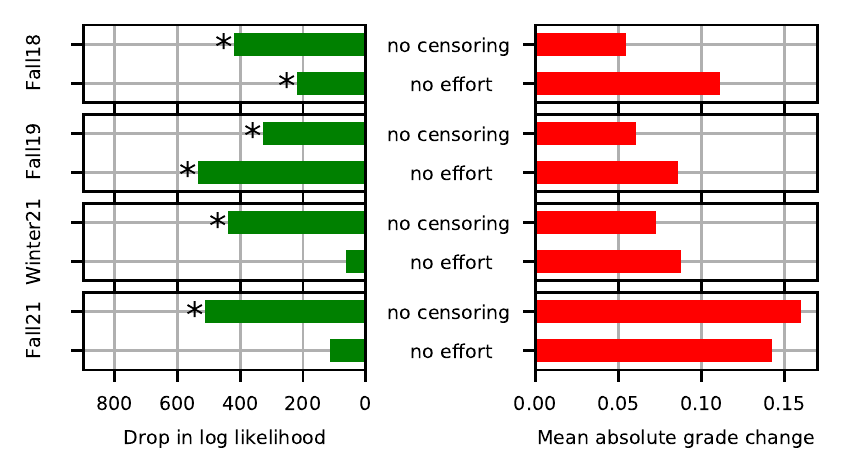} 
    \caption{Model ablations. Each bar represents the change relative to the best-performing model, which included both censoring and biases for each dataset. Stars on log-likelihood bars show significance ($p < 0.05$).}
    \label{fig:ablation}
\end{figure}

We then ran an ablation experiment, disabling each feature while holding the hyperparameters fixed.
We evaluated held-out likelihood of each ablated model along with the average absolute change in each student's grade. 
Our results, shown in Figure~\ref{fig:ablation}, indicate that both features improved the model's fit to data: removing censoring always caused significant drops in performance, while removing efforts had significant effects in the first two classes.
(The latter result is perhaps unsurprising: the later classes took greater lengths to disincentivize low-effort behavior, so detecting it is likely to have less impact on model performance.)
All model changes made small but meaningful changes to the final grades, averaging between 0.05 and 0.15 points on our 5-point scale, and seldom exceeding a single point change.

Finally, we also investigated variations in model architecture beyond the two extensions introduced in this paper. First, we asked whether we could get good performance without bias terms. We could not; they helped substantially. Second, we asked whether PG5-style correlation between student's grades on their own submissions and their reliabilities was helpful. It was not, regardless of whether we included biases. Details of both experiments are given in Appendix~\ref{sec: ablations_appendix}.

Having determined the superiority of our effort + censoring model, we used it to evaluate the extent to which the MIP had to deviate from submissions' MAP grades in order to present them as weighted averages of the graders' reports.
We set the MIP constants to the defaults recommended in Appendix~\ref{sec: mip}, allowing the graders' weights to change by at most $S=0.09$, with a minimum non-zero weight of $T=0.1$.
We found that replacing the MAP grades with the MIP's output would have caused only 3 to 6 percent of grades to change.


\section{Robustness Experiments on Synthetic Data}
\label{sec: synthetic_data}
While our experiments on real class data allowed us to test how well our models described real peer-grading behavior, they gave us no way to check the accuracy of the parameter estimates.
Of course, giving accurate grades (either for submissions or students' grading abilities) is a primary focus of peer grading systems.
We therefore conducted further experiments on \emph{synthetic} data, generating parameters and reported grades according to our best-fitting models from the previous section, and evaluating how well the posteriors recovered the latent parameters' true values.
This methodology allowed us to test how our estimates improved as the amount of grading data increased, how sensitive they were to the choice of hyperparameters, and whether explaining grades with our MIP increased their error.

The previous section showed that a model incorporating both efforts and censoring had the best performance on all four datasets, but its optimal hyperparameters varied, with only two datasets having exactly the same optimal hyperparameters.
Thus, we show results here based on one representative dataset, with full results for all three sets of optimal hyperparameters in Appendix~\ref{sec: parameter_recovery_appnedix} and~\ref{sec: misspec_appendix}. 
Unless otherwise specified, we simulated courses consisting of 10 weeks, with 120 students each making 1 submission and grading 4 peers' submissions each week.
The grading rubric had four components, each of which was given an integer grade between 0 to 5.
We also included 3 TAs who grade 25\% of submissions; we clamped their effort parameters to 1. We simulated TAs as being more reliable than most students: inspired by our real data, we gave TAs a mean reliability of 2 and gave students' mean reliability of 1.

We evaluated true grade and bias estimates via \emph{Mean Absolute Error (MAE)}. We also computed accuracy (the fraction of true grade MAP estimates equal to the rounded true grade) and RMSE, finding qualitatively similar conclusions with these measures.
We found that some models produced inaccurate reliability and effort estimates, but judged that this was less important because rewarding good grading only requires students to be \textit{ranked} in the correct order. Accordingly, we evaluated our reliability and effort estimates with the Spearman rank-order correlation coefficient, measuring how similarly students were ranked by the estimates and true values.
In each case, we report the mean and 95\% confidence intervals of each metric across inference runs on 15 simulated datasets.
We also compare our true grade MAEs to a hypothetical TA with a reliability of 2 (who achieves a mean absolute error of 0.48), allowing us to ask how much data is required to effectively substitute for a TA.

\subsection{Parameter Recovery}
\label{sec:parameter_recovery}
We begin by testing how the model's parameter estimates were affected by the amount of grading data available.
One obvious way to control the amount of data is to change the number of students in the simulated class. However, this change had surprisingly little impact on the inference problem's difficulty, because as the class size varies, each grader continued to grade a total of 40 submissions, and each submission continued to receive 4 grades.
Instead, we control these two dimensions separately, independently varying the number of grades from each student and the number of grades given to each submission. 

\begin{figure*}[t]
    \centering
    \includegraphics[width=0.75\linewidth,trim={0 0.4cm 0 0}]{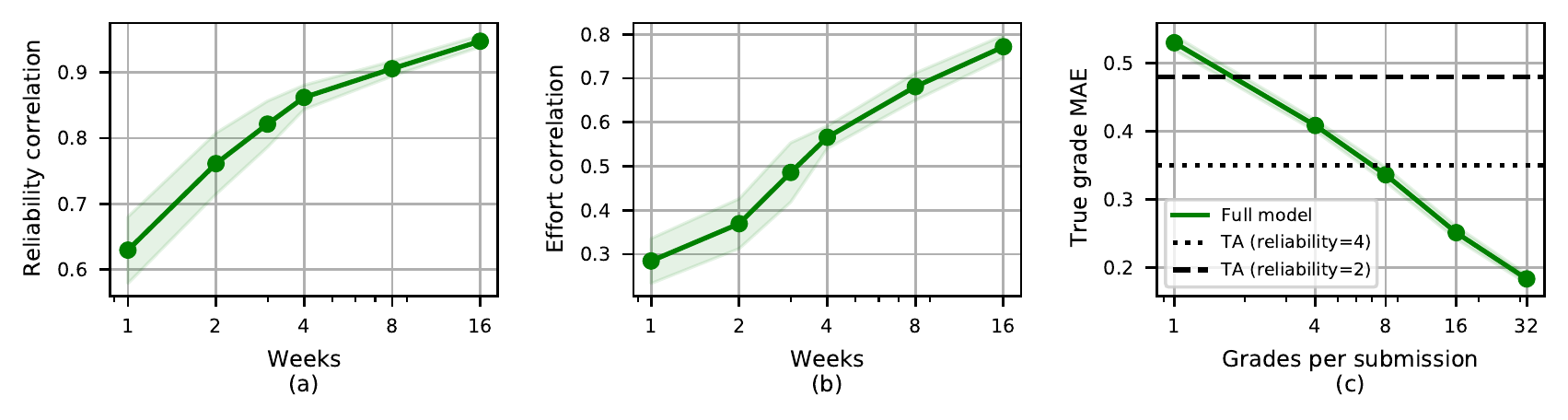} 
    \includegraphics[scale=0.75,trim={0 0.4cm 0 0}]{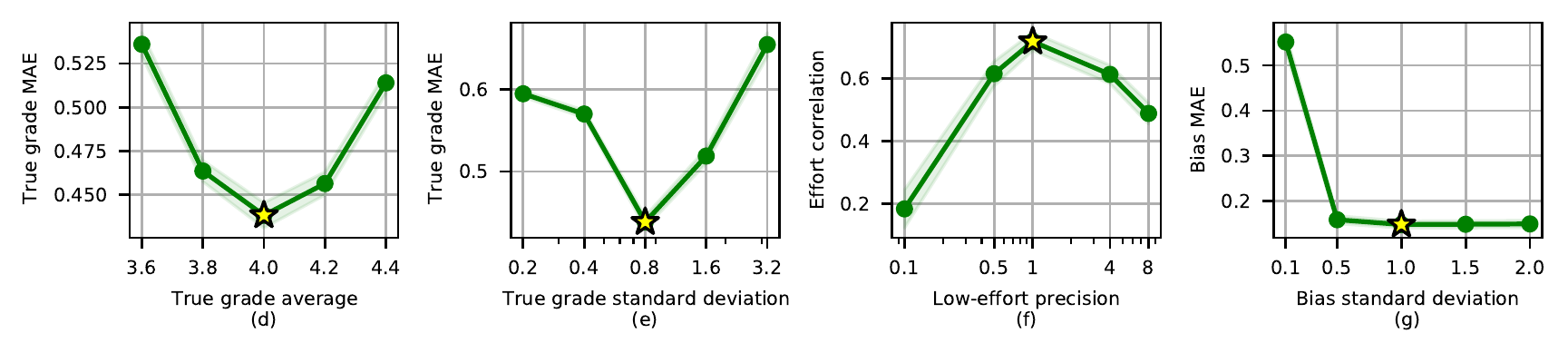}
    \caption{Effects of varying dataset size on (a) reliabilities; (b) effort probabilities; (c) true grades. Robustness of model outputs to misspecified hyperparameters: true grade (d) average and (e) stdev; (f) low-effort precision; (g) bias stdev.}
    \label{fig: consistency}
    \label{fig: misspec}
\end{figure*}

\paragraph{Varying grades per grader.}
First, we changed the number of grades given by each student by varying the number of weeks in the class.
Here, each assignment always received 4 grades, but the number of grades from each student scaled linearly with the number of weeks.
The results (Figure~\ref{fig: consistency}a-b) show that increasing the size of the dataset in this way improved grader quality estimates.
Reliability estimates had an appreciable correlation of 0.6 after just one week of data, improving substantially to 0.9 after 8 weeks.
Effort estimates followed a similar trend, but were much more difficult to estimate: one week of data produced a much poorer correlation of 0.3, with later weeks improving to 0.7.
Bias estimates, given in the appendix, also improved with additional data.

Perhaps surprisingly, true grade estimates improved very little as the number of weeks grew.
This suggests that, with only 4 grades per submission, most of the inaccuracy in the model's true grade estimates was driven by aggregating a small number of noisy signals, rather than because estimates of graders' reliabilities, biases, and efforts were inaccurate.

\paragraph{Varying grades per submission.} 
Next, we changed the number of graders in each course, holding the number of weekly submissions fixed at 40 and each student's grading workload at 4 grades per week. 
Adding graders in this way increased the number of grades given to each submission but preserved the amount of data about each grader, isolating the effect of additional information on each assignment. 
The impact of this change on true grade recovery is shown in Figure~\ref{fig: consistency}c.
These results indicate that adding additional peer graders on each submission substantially reduced true grade MAE, from 0.52 with a single grader to 0.41 with four---well below the MAE of a TA.
Adding more graders decreased true grade MAEs far lower, reaching below 0.2 with 32 graders. 

Increasing the amount of data in this way had little effect on the model's ability to recover students' reliabilities, biases, and effort probabilities.
This suggests that error in those estimates was driven primarily by noise in the reported grades, not by noise in the underlying true grade estimates.



\subsection{Robustness to Incorrect Hyperparameters}
\label{sec:synthetic_robustness}
While the synthetic experiments we have discussed so far show that we were able to recover the model's parameters with sufficient data, they assumed knowledge of the hyperparameters used to generate this data.
We now ask whether it is crucial to set these hyperparameters correctly, or whether the model still robustly recovers parameters of interest when given different hyperparameter settings than those used to generate the data.
We tested our models under seven changes to the hyperparameters, varying the true grade mean $\mu_s$; the true grade standard deviation $\sigma_s$; the bias standard deviation $\sigma_b$; the reliability prior mean $\alpha_\tau / \beta_\tau$; the reliability prior variance $\alpha_\tau / \beta_\tau^2$; the effort probability mean $\alpha_e / \beta_e$; and the low-effort precision $\tau_\ell$. 


Overall, we found that many of these changes to the hyperparameters had small and statistically insignificant effects on the inference results; these complete results are shown in Appendix~\ref{sec: misspec_appendix}.
Notably, we found that the model's performance was quite robust to changes in the reliability and effort probability priors. We show four exceptions in Figure~\ref{fig: misspec}: using an incorrect mean or standard deviation for the true grade prior substantially increased true grade MAE from 0.44 to as high as 0.53; incorrectly specifying the low effort distribution $\tau_\ell$ was very detrimental to the effort probability estimates; and using a bias prior with a standard deviation far below its true value hurt bias estimates.

\subsection{MIP stability}

\begin{figure}[t]
    \centering
    \includegraphics[width=0.8\linewidth,trim={0 0.5cm 0 0}]{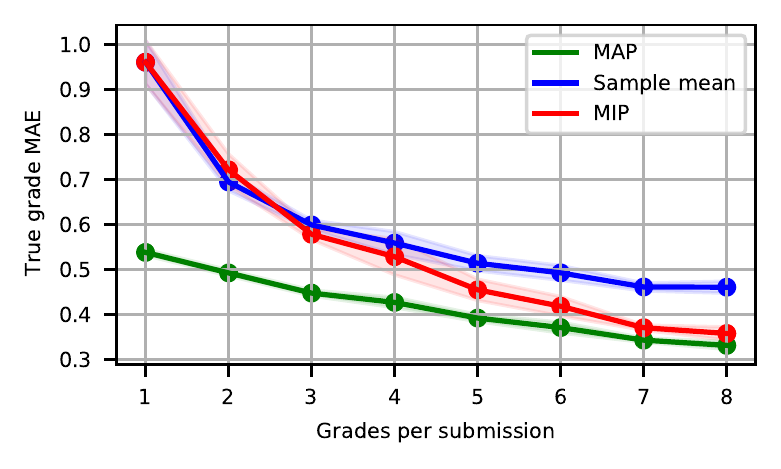} 
    \caption{Effect of MIP explanations on true grade MAE.}
    \label{fig: mip_stability_synthetic}
\end{figure}

Lastly, we tested the impact of assigning grades based on explanations from our MIP formulation, rather than our model's MAP estimates of true grades, as we varied the number of peer grades given to each submission.
The results are shown in Figure~\ref{fig: mip_stability_synthetic}.
With only a single grader, the MIP output was equal to the mean of the Gibbs samples, which is much less accurate than the MAP. 
However, with additional graders, the MIP gained the flexibility to recover MAP grades, reaching nearly equal MAEs at 7 graders per submission.


\section{Conclusions and Practical Considerations}
\label{sec: discussion}
We have shown how probabilistic peer grading systems can be extended to provide incentives for effortful grading; to correctly model discrete peer grades; and to output discrete, interpretable final grades that approximate MAP estimates. We validated our models on four years of real classroom data and investigated both their ability to recover true parameters and their robustness on synthetic data. 

Although the peer grading literature has repeatedly shown that Bayesian models can produce accurate grades, tuning them to produce such good performance can be a daunting task for an instructor---our model has 9 hyperparameters! Luckily, our robustness experiments in Section~\ref{sec:synthetic_robustness} showed that the model's posterior beliefs were robust to misspecifying the reliability, bias, and effort priors.
Two hyperparameters remain. The first is the true grade distribution, a choice that instructors often make when curving grades. The second is the specification of low-effort behavior, which is important both for boosting model performance and for disincentivizing bad behavior: if the model is good at identifying low effort behavior, students will exhibit this behavior less often. 
We recommend adapting the specification to capture low-effort behavior observed in spot checks. 


Our insistence on providing uncertainty estimates is not just a statistical concern. 
Our methods work best when they are integrated into the design of the class, giving these uncertainty estimates pedagogical value.
For assignment grades, uncertainty estimates can direct TA spot checks towards areas of disagreement. 
For grader reliability, uncertainty estimates can inform whether students should be trusted to peer grade without TA supervision.
They can also help evaluate students' peer grading prowess: in our own class, we initialized the model to be confident that students had poor reliability and required students to do extra grading if the model's pessimistic estimate of their reliability was poor, but scored their peer grading based on the model's \textit{optimistic} reliability estimate. 
Thus, students got the best grade the model could justify, but students suspected to be weak got additional practice grading, which refined our reliability estimates in turn.

Our parameter recovery experiments in Section~\ref{sec:parameter_recovery} found that graders' effort probabilities were difficult to estimate: compared to reliabilities, effort probability estimates were much poorer with little data, and converged more slowly as data became available.
This is not a surprise: our low-effort graders choose grades that are as difficult as possible to distinguish from effortful graders.
The problem is exacerbated by coarse rubrics, which cause many high-effort grades to match the class average exactly.
Performance could be improved by tuning the specification of low effort behavior, using an autograding system as another unbiased signal about submissions' grades~\cite{han2020human}, or by leveraging other signals of low-effort behavior, such as graders' time spent grading and typing speed~\cite{wang2019improving}.

\section*{Acknowledgments}
This work was funded by an NSERC Discovery Grant, a DND/NSERC Discovery Grant Supplement, a CIFAR Canada AI Research Chair (Alberta Machine Intelligence Institute), a Compute Canada RAC Allocation, awards from Facebook Research and Amazon Research, and DARPA award FA8750-19-2-0222, CFDA \#12.910 (Air Force Research Laboratory).

\bibliographystyle{ACM-Reference-Format}
\bibliography{references}


\clearpage
\appendix
\setcounter{page}{1}

\section{PG5: Correlating Submissions and Graders}
\label{sec: PG5}

PG1 ignores the identities of each submission's authors. 
In practice, though, students' grades and grading abilities tend to correlate through their underlying abilities, time commitment, and interest in the course.
Piech et al's PG3 captured this correlation, which models graders' reliabilities as depending on their own submissions' grades in a deterministic way.
\citet{mi2015probabilistic} took this idea further with PG5, assuming that each grader's reliability is non-deterministically influenced by the true grade of their own submission---that is, that students with higher grades are often, but not always, better graders. 

\subsection{Revisiting PG5's Correlation Structure}
The PG5 model is defined only for classes with a single component. 
Abusing notation, let $s_u$ be the true grade of submission $u$ on its only component, and let $v(u)$ be the grader that authored submission $u$.
Then, the PG5 data generating process is:
\begin{alignat*}{2}
    \text{(True grades)}& \quad 
    s_u &&\sim \NormalDist(\mu_s, 1 / \tau_s); \\
    \text{(Reliabilities)}& \quad 
    \tau^{v(u)} &&\sim \NormalDist(s_u, 1 / \beta_0); \\
    \text{(Biases)}& \quad 
    b^v &&\sim \NormalDist(0, 1 / \tau_b); \\
    \text{(Peer grades)}& \quad
    g_u^v &&\sim \NormalDist(s_u + b^v, \lambda / \tau_v).
\end{alignat*}
Introducing this correlation structure has a computational drawback: while the true grade and bias Gibbs updates still have closed forms, the reliability update does not.
To get around this problem, Mi and Yeung use a discrete approximation of the reliability Gibbs update: they compute an unnormalized posterior probability for many potential values of a student's reliability, normalize these probabilities to sum to 1, and sample from the resulting distribution.
Despite this computational difficulty, they show that this model better recovers staff grades on real peer grading data.

While it is possible to extend this 
correlation structure to classes with multiple assignments or rubric elements---for example, by conditioning a student's reliability on the average of all of their submissions' true grades---the resulting model has a key flaw.
Because this model has an arrow from each true grade to the grader's reliability (a V-structure), observing a high true grade on one submission or rubric element increases the probability of \emph{lower} grades on the others.

\begin{figure}
    \centering
    \begin{tabular}{cc}
        \includegraphics[width=0.45\textwidth]{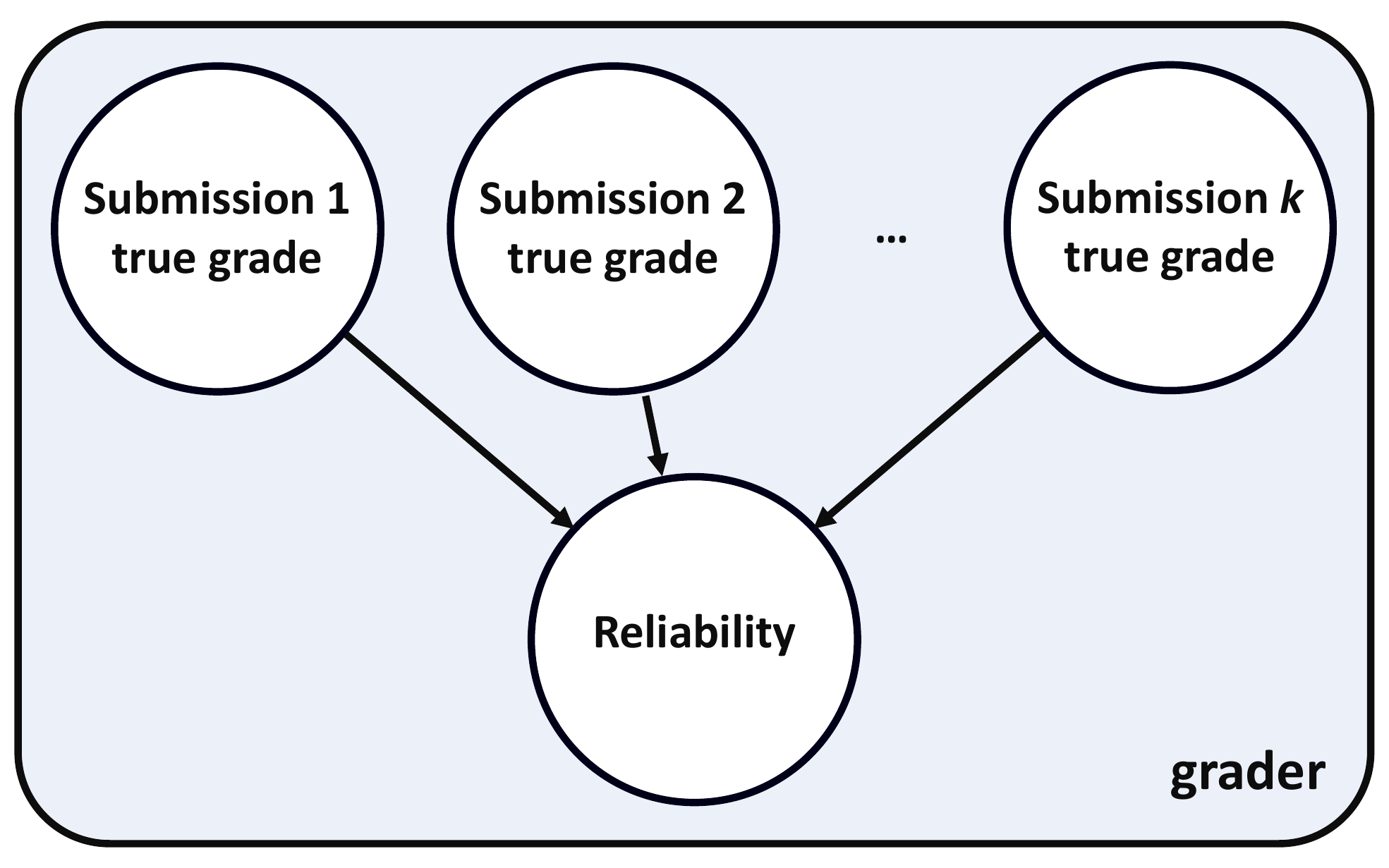} &
        \includegraphics[width=0.45\textwidth]{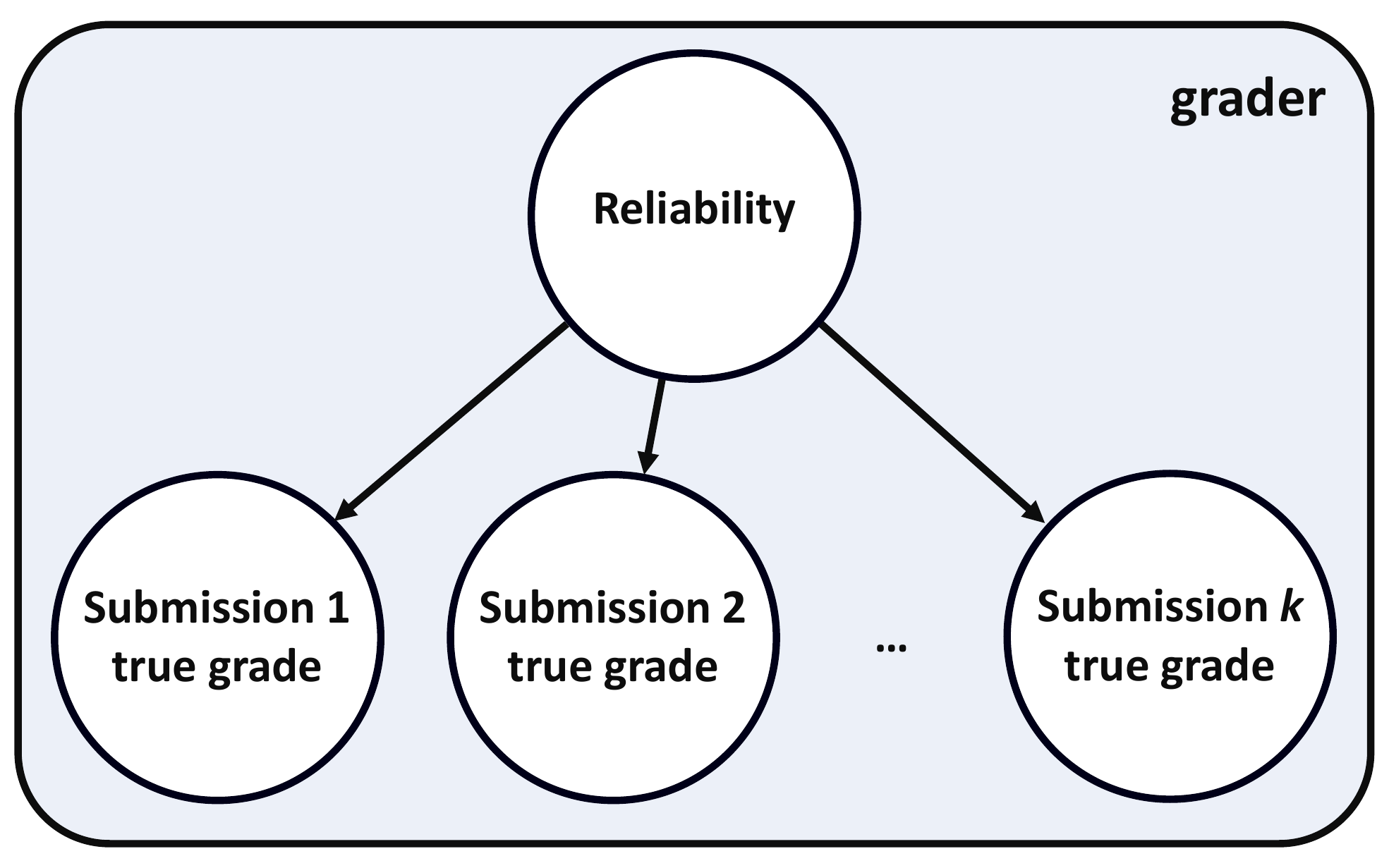} \\
        (a) &
        (b)
    \end{tabular}
    \caption{(a) PG5's model of correlation between true grades and reliabilities. Conditioned on a student's reliability, their true grades are negatively correlated. (b) Our updated model. Conditioned on a student's reliability, their true grades are independent.}
    \label{fig:pg5}
\end{figure}

We argue that these arrows should be flipped, as in Figure~\ref{fig:pg5}:
true grades should be modeled as conditionally dependent on their author's reliability rather than the reverse. 
This model still recovers PG5 in the single-submission, single-component case.  
Formally, our complete model is
\begin{alignat*}{2}
    \text{(Reliabilities)}& \quad 
    \tau^v &&\sim \NormalDist(\mu_s, 1 / \tau_s + 1 / \beta_0); \\
    \text{(True grades)}& \quad 
    s_{u,c} &&\sim \NormalDist\left(\frac{\tau_s \mu_s + \beta_0 \tau^{u(v)}}{\tau_s + \beta_0}, \frac{1}{\tau_s + \beta_0}\right); \\
    \text{(Biases)}& \quad 
    b^v &&\sim \NormalDist(0, 1/\tau_b); \\
    \text{(Effort probabilities)}& \quad 
    e^v &&\sim \BetaDist(\alpha_e, \beta_e); \\
    \text{(Efforts)}& \quad
    z_u^v &&\sim \Ber(e^v); \\
    \text{(Peer grades)}& \quad 
    g_{u,c}^v &&\sim \begin{cases}
        \NormalDist\left(s_{u,c} + b^v, \lambda / \tau^v\right) & z_u^v = 1; \\
        D_\ell, & z_u^v = 0.
    \end{cases}
\end{alignat*}
We also consider a censored version of this model, which adds reported grade variables that round each peer grade.

We include this feature in our experiments on real peer grading data in Appendix~\ref{sec: hyperparam_search} and \ref{sec: ablations_appendix}, where we show that correlation sometimes significantly hurts held-out likelihoods and is never a necessary feature of the best model. 

\subsection{Gibbs Updates (Without Censoring)}
Next, we provide Gibbs updates for the uncensored version of this model. 
First, notice that adding correlation between true grades and reliabilities has no effect on the bias, effort probability, and effort updates: because these updates condition on values of the reliability variables, the process by which these values are generated is irrelevant to the update rules.
The true grade update also has a closed form:
\begin{alignat*}{2}
    \text{(True grades)} &\quad
    \hat s_{u,c} &&\sim \NormalDist \left(
        \frac{\tau_s \mu_s + \beta_0 \hat \tau^{u(v)} + \sum_{v \in \matching_u} \hat \tau^v (g_{u,c}^v - \hat b^v)}{\tau_s + \beta_0 + \sum_{v \in \matching_u} \hat \tau^v},
        \frac{1}{\tau_s + \beta_0 + \sum_{v \in \matching_u} \hat \tau^v}
    \right).
\end{alignat*}
The reliability update, on the other hand, does not have a simple closed form.
Instead, as in our Gibbs updates for censored models, we make a discrete approximation, using the unnormalized posterior
\begin{align*}
    \text{(Reliabilities)} \quad
    \Pr(\tau^v | \bm{\hat s}, \bm{\hat b}, \bm{\hat z}, \bm{r})
    \propto 
    &\Pr(\tau^v | \mu_s, \tau_s, \beta_0)
    \cdot \prod_{u: v(u) = v} \Pr(\hat s_{u,c} | \tau^v, \mu_s, \tau_s, \beta_0) \\
    &\cdot \prod_{\substack{u \in \matching_v \\ z_u^v = 1}} \prod_{c=1}^C \varphi(g_{u,c}^v | \hat s_{u,c} + \hat b^v, 1 / \tau^v).
\end{align*}

\subsection{Gibbs Updates (With Censoring)}
Lastly, we add correlation to the fully featured model (with efforts and censoring).
As with the uncensored model, adding correlation has no impact on the bias, effort probability, or effort updates.
The remaining two, which still require discrete approximations, have the unnormalized posteriors
\begin{alignat*}{2}
    \text{(True grades)} &\quad
    \Pr(s_{u,c} | \bm{\hat \tau}, \bm{\hat b}, \bm{\hat z}, \bm{r}) 
    &&\propto \Pr(s_{u,c} | \hat \tau^{v(u)}, \mu_s, \tau_s, \beta_0) \prod_{v \in \matching_u} L(r_{u,c}^v | s_{u,c}, \hat \tau^v, \hat b^v, \hat z_u^v); \\
    \text{(Reliabilities)} &\quad
    \Pr(\tau^v | \bm{\hat s}, \bm{\hat b}, \bm{\hat z}, \bm{r})
    &&\propto \Pr(\tau^v | \mu_s, \tau_s, \beta_0)
    \cdot \prod_{u: v(u) = v} \Pr(\hat s_{u,c} | \tau^v, \mu_s, \tau_s, \beta_0) \\
    & &&\quad \cdot \prod_{u \in \matching_v} \prod_{c=1}^C L(r_{u,c}^v | \hat s_{u,c}, \hat \tau^v, b^v, \hat z_u^v)
\end{alignat*}

\clearpage

\section{Gibbs Updates}
\label{sec: gibbs_updates}
In this section, we list Gibbs updates for the models described in the paper.
We write $\matching \subseteq \submissions \times \graders$ to denote the set of available peer grades, where $(u, v) \in \matching$ indicates that submission $u \in \submissions$ was graded by grader $v \in \graders$.
We write $\matching_u = \{ v: (u, v) \in \matching\}$ to denote the set of graders that graded a submission $u$, and $\matching_v = \{ u: (u, v) \in \matching\}$ for the set of submissions graded by a single grader $v$. 
Boldface symbols denote the sets of variables: for example, $\bm{s} = \{s_{u,c}: u \in \submissions, c \in \{1, \dots, C\}\}$ is the set of all true grades.
We also denote the pdf and cdf of the normal distribution $\mathcal{N}(\mu, \sigma^2)$ as 
\[
    \varphi(x | \mu, \sigma^2) = \frac{1}{\sigma \sqrt{{2\pi}}} \exp\left(-\frac{(x - \mu)^2}{2\sigma^2} \right)
\]  
and $\Phi(x | \mu, \sigma^2) = \int_{-\infty}^x \varphi(t | \mu, \sigma^2) dt$.

\subsection{PG1}

First, we restate the Gibbs updates for PG1 that were originally derived by~\citet{piech2013tuned}.
Recall that the PG1 model is:
\begin{alignat*}{2}
    \text{(True grades)}& \quad 
    s_{u,c} &&\sim \mathcal{N}(\mu_s, 1 / \tau_s); \\
    \text{(Reliabilities)}& \quad 
    \tau^v &&\sim \GammaDist(\alpha_\tau, \beta_\tau); \\
    \text{(Biases)}& \quad 
    b^v &&\sim \mathcal{N}(0, 1 / \tau_b); \\
    \text{(Peer grades)}& \quad
    g_{u,c}^v &&\sim \mathcal{N}(s_{u,c} + b^v, 1 / \tau^v).
\end{alignat*}

Thanks to the conjugate structure of this model, given a current belief about the true grades $\bm{\hat s}$, reliabilities $\bm{\hat \tau}$, and biases $\bm{\hat b}$, each variable's update has a simple closed form:
\begin{alignat*}{2}
	\text{(True grades}) & \quad
        \hat s_{u,c} &&\sim \mathcal{N} \left( %
    	\frac{\tau_s \mu_s + \sum_{v \in \matching_u} \hat \tau^v (g_{u,c}^v - \hat b^v)}{\tau_s + \sum_{v \in \matching_u} \hat \tau^v}, %
    	\frac{1}{\tau_s + \sum_{v \in \matching_u} \hat \tau^v} %
	\right); \\
        \text{(Reliabilities)}& \quad 
	\hat \tau^v &&\sim \GammaDist \left( %
    	\alpha_\tau + \frac{C}{2} |\matching_v|,
            \beta_\tau + \frac{1}{2} \sum_{u \in \matching_v} \sum_{c=1}^C \left(g_{u,c}^v - \hat b^v - \hat s_{u,c}\right)^2 %
	\right); \\
        \text{(Biases)}& \quad 
	\hat b^v &&\sim \mathcal{N} \left( %
    	\frac{\sum_{u \in \matching_v} \hat \tau^v (g_{u,c}^v - \hat s_{u,c})}{\tau_b + \sum_{u \in \matching_v} \hat \tau^v }, %
            \frac{1}{\tau_b + \sum_{u \in \matching_v} \hat \tau^v} %
	\right).
\end{alignat*}

\subsection{Adding Effort}

Next, we show how the PG1 Gibbs updates can be adapted to include our model of student effort.
Including variables for peer graders' efforts and effort probabilities, the model becomes:
\begin{alignat*}{2}
    \text{(True grades)}& \quad 
    s_{u,c} &&\sim \mathcal{N}(\mu_s, 1 / \tau_s); \\
    \text{(Reliabilities)}& \quad 
    \tau^v &&\sim \GammaDist(\alpha_\tau, \beta_\tau); \\
    \text{(Biases)}& \quad 
    b^v &&\sim \mathcal{N}(0, 1 / \tau_b); \\
    \text{(Effort probabilities)}& \quad 
    e^v &&\sim \BetaDist(\alpha_e, \beta_e); \\
    \text{(Efforts)}& \quad
    z_u^v &&\sim \Ber(e^v); \\
    \text{(Peer grades)}& \quad
    g_{u,c}^v &&\sim \begin{cases}
        \NormalDist(s_{u,c} + b^v, 1 / \tau^v), & z_u^v = 1; \\
        D_\ell, & z_u^v = 0.
    \end{cases}
\end{alignat*}
This more complex model continues to enjoy closed-form Gibbs updates.
Let $p_\ell(g)$ and $P_\ell(g)$ denote the pdf and cdf of $D_\ell$, respectively, at the grade $g$.
Then, the updates are
\begin{alignat*}{2}
        \text{(True grades)}& \quad 
	\hat s_{u,c} &&\sim \mathcal{N} \left( 
    	\frac{\tau_s \mu_s + \sum_{v \in \matching_u} \hat z_u^v \hat \tau^v (g_{u,c}^v - \hat b^v)}{\tau_s + \sum_{v \in \matching_u} \hat z_u^v \hat \tau^v}, 
    	\frac{1}{\tau_s + \sum_{v \in \matching_u} \hat z_u^v \hat \tau^v}
	\right); \\
        \text{(Reliabilities)}& \quad 
	\hat \tau^v &&\sim \GammaDist \left( %
    	\alpha_\tau + \frac{C}{2} \sum_{u \in \matching_v} \hat z_u^v, 
            \beta_\tau + \frac{1}{2} \sum_{u \in \matching_v} \sum_{c=1}^C \hat z_u^v (g_{u,c}^v - b^v - \hat s_{u,c})^2 %
	\right); \\    
        \text{(Biases)}& \quad 
	\hat b^v &&\sim \mathcal{N} \left( %
    	\frac{\sum_{u \in \matching_v} \hat \tau^v (g_{u,c}^v - \hat s_{u,c})}{\tau_b + \sum_{u \in \matching_v} \hat \tau^v }, %
            \frac{1}{\tau_b + \sum_{u \in \matching_v} \hat \tau^v} %
	\right); \\
        \text{(Effort probabilities)}& \quad
	e^v &&\sim \BetaDist\left(
            \alpha_e + \sum_{u \in \matching_v} \hat z_u^v, 
    	\beta_e + \sum_{u \in \matching_v} (1 - \hat z_u^v)
	\right); \\
        \text{(Efforts)}& \quad
	\hat z_u^v &&\sim \Ber \left( %
    	\frac{\hat e^v \prod_{c=1}^C \varphi(g_{u,c}^v | \hat s_{u,c} + \hat b^v, 1 / \hat \tau^v)}
    	{\hat e^v \prod_{c=1}^C \varphi(g_{u,c}^v | \hat s_{u,c} + \hat b^v, 1 / \hat \tau^v) + (1 - \hat e^v) \prod_{c=1}^C p_\ell(g_{u,c}^v)}
        \right).
\end{alignat*}
Notice that the true grade, reliability, and bias updates are very similar to the PG1 updates: the difference is that they ignore any peer grades that are believed to be low-effort, recovering the original PG1 updates if all $z_u^v$ are set to $1$.

\subsection{Adding Censoring}
Lastly, we add censoring, treating the peer grades as latent variables and observing rounded reported grades instead.
The complete model is:
\begin{alignat*}{2}
    \text{(True grades)}& \quad 
    s_{u,c} &&\sim \mathcal{N}(\mu_s, 1 / \tau_s); \\
    \text{(Reliabilities)}& \quad 
    \tau^v &&\sim \GammaDist(\alpha_\tau, \beta_\tau); \\
    \text{(Biases)}& \quad 
    b^v &&\sim \mathcal{N}(0, 1 / \tau_b); \\
    \text{(Effort probabilities)}& \quad 
    e^v &&\sim \BetaDist(\alpha_e, \beta_e); \\
    \text{(Efforts)}& \quad
    z_u^v &&\sim \Ber(e^v); \\
    \text{(Peer grades)}& \quad
    g_{u,c}^v &&\sim \begin{cases}
        \NormalDist(s_{u,c} + b^v, 1 / \tau^v), & z_u^v = 1; \\
        D_\ell, & z_u^v = 0;
    \end{cases} \\
    \text{(Reported grades)}& \quad
    r_{u,c}^v &&= n_G(g^v_{u,c}).
\end{alignat*}

As described in the main text, sampling the peer grades $g_{u,c}^v$ leads to a highly sample-inefficient inference algorithm, as the resulting posterior is highly multimodal.
Instead, we treat the peer grades as nuisance parameters, integrating over all of their possible values in each update.

The effort probability update rule is unaffected by this change to the model, as effort probabilities are independent of peer grades and reported grades conditional on efforts.
For the remaining updates, it is useful to introduce some additional notation.
First, let $\underline{g}(r)$ and $\bar{g}(r)$ be the minimum and maximum peer grades that are reported as $r$, respectively.
Then, for a fixed true grade $s_{u,c}$, reliability $\tau^v$, and effort $z_u^v$, the likelihood of a reported grade $r_{u,c}^v$ is
\begin{align*}
    L(r_{u,c}^v | s_{u,c}, \tau^v, b^v, z_u^v)
    &= \int_{\underline{g}(r_{u,c}^v)}^{\bar{g}(r_{u,c}^v)} \Pr(g_{u,c}^v | s_{u,c}, \tau^v, b^v, z_u^v) dg_{u,c}^v \\
    &= \begin{cases}
        \Phi\left(\bar{g}(r_{u,c}^v) | s_u + b^v, 1/\tau^v\right) - \Phi\left(\underline{g}(r_{u,c}^v) | s_u + b^v, 1/\tau^v\right), &z_u^v = 1; \\
        P_\ell\left(\bar{g}(r_{u,c}^v)\right) - P_\ell\left((\underline{g}(r_{u,c}^v)\right), &z_u^v = 0.
    \end{cases}
\end{align*}
Using this notation, the Gibbs updates for efforts can be written in closed form:
\begin{alignat*}{2}
    \text{(Efforts)}& \quad 
    \hat z_u^v &&\sim \Ber\left(
        \frac{\hat e^v \prod_{c=1}^C L(r_{u,c}^v | \hat s_{u,c}, \hat \tau^v, \hat b^v, 1)}
        {\hat e^v \prod_{c=1}^C L(r_{u,c}^v | \hat s_{u,c}, \hat \tau^v, \hat b^v, 1) + (1 - \hat e^v) \prod_{c=1}^C L(r_{u,c}^v | \hat s_{u,c}, \hat \tau^v, \hat b^v, 0)}
    \right)
\end{alignat*}

Unlike the previous models, the true grade, reliability, and bias updates do not have closed forms after adding censoring.
Instead, we use a discrete grid-based approximation for these updates.
We update each of these variables by considering a uniform grid of possible values, calculating an unnormalized posterior probability for each value, renormalizing these probabilities to sum to 1, and taking a sample from the resulting discrete distribution.
These unnormalized posterior probabilities are:
\begin{alignat*}{2}
    \text{(True grades)} &\quad
    \Pr(s_{u,c} | \bm{\hat \tau}, \bm{\hat b}, \bm{\hat z}, \bm{r}) 
    &&\propto \Pr(s_{u,c} | \mu_s, \tau_s) \prod_{v \in \matching_u} L(r_{u,c}^v | s_{u,c}, \hat \tau^v, \hat b^v, \hat z_u^v); \\
    \text{(Reliabilities)} &\quad
    \Pr(\tau^v | \bm{\hat s}, \bm{\hat b}, \bm{\hat z}, \bm{r})
    &&\propto \Pr(\tau^v | \alpha_\tau, \beta_\tau) \prod_{u \in \matching_v} \prod_{c=1}^C L(r_{u,c}^v | \hat s_{u,c}, \tau^v, \hat b^v, \hat z_u^v); \\
    \text{(Biases)} &\quad
    \Pr(b^v | \bm{\hat s}, \bm{\hat \tau}, \bm{\hat z}, \bm{r})
    &&\propto \Pr(b^v | \tau_b) \prod_{u \in \matching_v} \prod_{c=1}^C L(r_{u,c}^v | \hat s_{u,c}, \hat \tau^v, b^v, \hat z_u^v)
\end{alignat*}

In practice, with integer-valued reported grades ranging from $0$ to $5$, we use the following grids:
\begin{itemize}
    \item True grades: 101 values uniformly spaced from 0 to 6;
    \item Reliabilities: 100 values uniformly spaced from 0.1 to 10;
    \item Biases: 61 values uniformly spaced from -3 to 3.
\end{itemize}

\clearpage

\section{MIP Formulation}
\label{sec: mip}
Fix a submission $u$. 
For each component $c$, let $g_{u,c} \in \{1, \dots, M\}$ be the final grade, and let $g_{u,c:k} \in \{0, 1\}$ be a one-hot vector where $g_{u,c:k}=1$ if and only if $g_{u,c} = k$.
Let $m_{u,c:k}$ be the fraction of Gibbs samples for $s_{u,c}$ that round to $k$ (i.e., are within in the interval $[k-0.5, k+0.5]$).
Let $d_u^v \in \mathds{R}^+$ be a set of desired weights for each grader $v$ on this submission.
Then, our MIP formulation is:
\begin{equation}
    \max \sum_{c,k} m_{u,c:k} \cdot g_{u,c:k} - P \sum_v (p_u^v + n_u^v)
\end{equation}
subject to:
\begin{align}
    &      \sum_k g_{u,c:k} = 1      \ \      \forall u \in U \text{ and }  \forall c \in [C],   \\
    &    g_{u,c} = \sum_k k \cdot g_{u,c:k} \ \   \forall u \in U \text{ and }  \forall c \in [C],  \ \      \\
    &    g_{u,c}  = \sum_v w_u^v \cdot r_{u,c}^v + s_{u,c}  \ \ \forall u \in U \text{ and }  \forall c \in [C],  \label{4}\\ 
     &   w_u^v = d_u^v + (p_u^v - n_u^v)     \ \   \forall  v \in V \text{ and }  \forall u \in U, \ \  \label{5}\\
     &    T \cdot e_u^v \leq w_u^v \leq e_u^v \ \          \forall  v \in V  \text{ and }  \forall u \in U,    \ \    \label{6}\\
     &   \sum_v w_u^v = 1 \ \  \forall u \in U, \\
     &   0 \leq w_u^v \leq 1  \ \ \forall v \in V \text{ and }  \forall u \in U, \\
     &    g_{u,c} \in \{1,2,\dots,M\} \ \ \forall c \in [C], \\
     &   g_{u,c:k} \in \{0, 1\} \ \ \forall u \in U, c \in [C], \text{ and }  \forall k \in  \{1,2,\dots,M\}, \\
    &    -0.5 \leq s_{u,c} \leq 0.5  \ \  \forall u \in U \text{ and }  \forall c \in [C],   \\
    &    0 \leq p_u^v, n_u^v \leq S  \ \ \forall  v \in V \text{ and }  \forall u \in U \label{eqn:max_weight_change}, \\
    &    e_u^v \in \{0, 1\} \ \ \forall  v \in V \text{ and }  \forall u \in U.
\end{align}
Notice in this formulation that \eqref{4} ensures that the grade for each component $c$ is a rounded weighted average of the reported grades; \eqref{5} imposes penalties for selecting weights that deviate from the desired weights; \eqref{6} ensures that no weights may be in the interval $(0, T)$; and \eqref{eqn:max_weight_change} guarantees that $|w_u^v - d_u^v| \le S$. 
In practice, we allow graders' weights to change by at most $S = 0.09$, allow a minimum non-zero weight of $T = 0.1$, and set the penalty term to $p = 0.01$.

While mixed integer programs can be computationally hard to solve in general, typical instances of our MIP with 4 peer grades were solved in a fraction of a second with an open-source MIP solver.
\clearpage

\section{Hyperparameter Search}
\label{sec: hyperparam_search}
Including the possibility of models with correlation (Appendix~\ref{sec: PG5}), our model family has 11 distinct hyperparameters. 
With such a large space of possible settings of their values, searching all combinations of these hyperparameters is computationally intractable. 
To cut down the difficulty of this optimization problem, we began with exploratory experiments, pruning hyperparameter values that resulted in relatively poor performance. 
Specifically, we initially set the model's hyperparameters to values that we found to behave well in a real class setting (e.g., showed meaningful variation in students' effort probabilities, didn't assign enormous reliabilities to few students, etc.) and varied one hyperparameter at a time around this initial point.
During these exploratory experiments, we tested the values:
\begin{align*}
    \mu_s: &~ \{3, 3.5, 4, 4.5\}, \\
    \sigma_s: &~ \{0.1, 0.2, 0.5, 0.8, 1, 1.6, 2, 3\}, \\
    \sigma_b: &~ \{0, 0.1, 0.5, 1, 1.5, 2 \}, \\
    (\alpha_e, \beta_e): &~ \{(2,8), (5,5),(8,2),(9.5,0.5),(9.9,0.1) \}, \\
    (\alpha_\tau, \beta_\tau): &~ \{(1,2), (1,1), (2,2),(2,1), (3,1)\}, \\
    \sigma_\tau: &~ \{0.1, 0.5, 1, 2 , 3 , 4 , 5, 6 , 8 \}, \\
    \lambda_\tau: &~ \{0.5, 1, 2 , 3 , 4 \}, \\
    \tau_\ell: &~ \{0.5, 1 , 2, 3, 4, 6, 8\}, \\
    \epsilon: &~ \{0, 0.05, 0.1, 0.5\}.
\end{align*}

We found that many of these settings produced models with substantially worse held-out likelihoods on our class data. 
Based on these findings, we defined a more tractable grid of parameters:
\begin{align*}
    \mu_s: &~ \{3.5, 4\}, \\
    \sigma_s: &~ \{0.8,  1.6\}, \\
    \sigma_b: &~ \{0.1, 0.5, 1 \}, \\
    (\alpha_e, \beta_e): &~ \{(5,5),(8,2)\}, \\
    (\alpha_\tau, \beta_\tau): &~ \{(1,1), (2,2),(2,1)\}, \\
    \sigma_\tau: &~ \{0.1, 2, 8 \}, \\
    \lambda_\tau: &~ \{1, 2, 4 \}, \\
    \tau_\ell: &~ \{ 1, 4\}, \\
    \epsilon: &~ 0.05.
\end{align*}
For the simpler models, this simplfication was enough for us to perform a grid search for the optimal hyperparameters.
(For example, in PG1, only five of these hyperparameters---$\mu_s$, $\sigma_s$, $\sigma_b$, $\alpha_\tau$, and $\beta_\tau$---have an effect on the model, resulting in a grid with only 36 combinations.)
For the most complex models, the resulting grid was still too large to test exhaustively, so we randomly sampled combinations of these parameters. 
In Section~\ref{sec: real_data} and Appendix~\ref{sec: ablations_appendix}, our reported held-out likelihoods refer to the best performance that we found searching this reduced grid.
In total, we ran 10-fold cross validation on 4943 models, taking a total of approximately 45 CPU-years of computation.

\clearpage 

\section{Additional Ablations}
\label{sec: ablations_appendix}

\begin{figure}
    \centering
    \includegraphics[trim={0 0.4cm 0 0}]{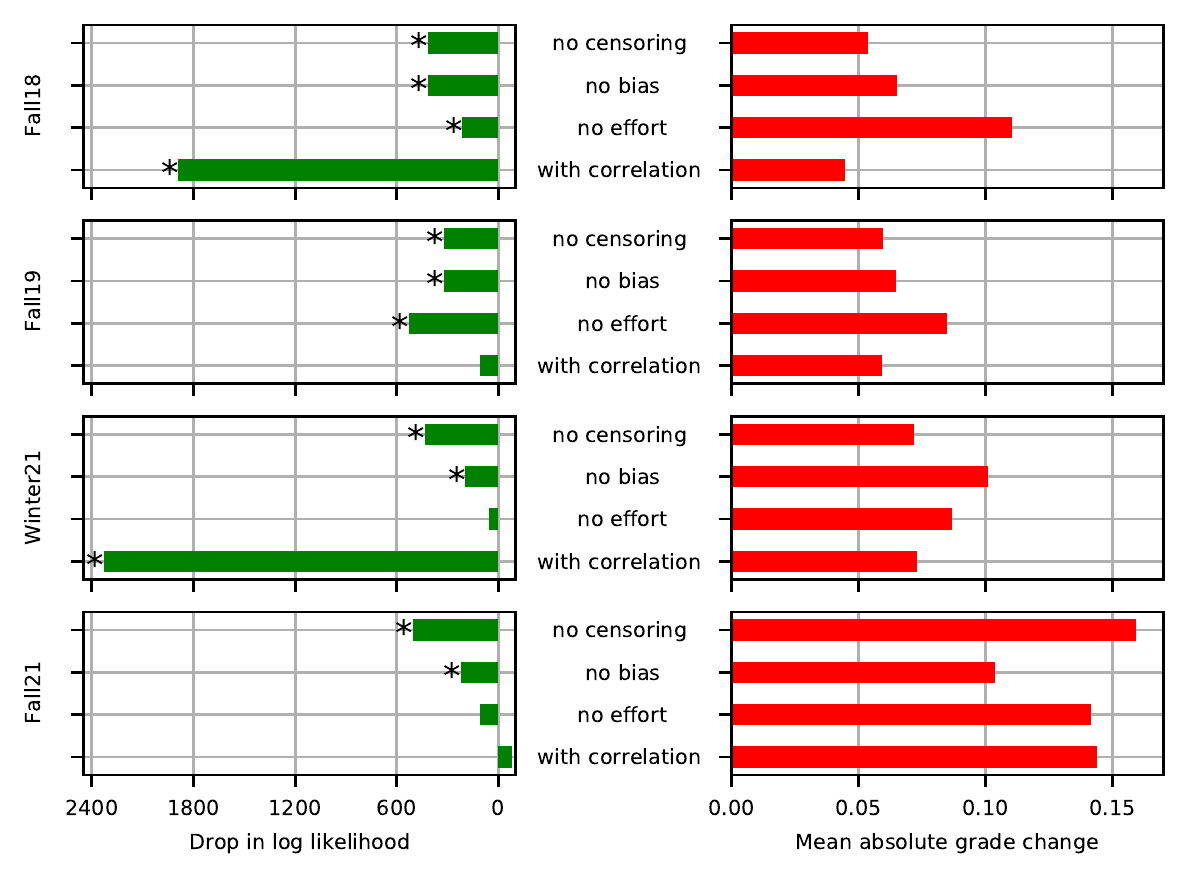} 
    \caption{Model ablations, including biases and correlation. Each bar represents the change relative to the best-performing model, which included both censoring and biases for each dataset. Stars on log-likelihood bars show significance ($p < 0.05$).}
    \label{fig: ablation_full}
\end{figure}

In this section, we include additional ablations on the best-performing model. 
We considered two additional features that we excluded from Section~\ref{sec: real_data}: the possibility of disabling biases, and the possibility of correlation between students' grades on their own submissions and their reliabilities (Appendix~\ref{sec: PG5}). 
Testing all combinations of these four features, we found that a model using biases, effort, and censoring, but not correlation, was the best model on three of our four datasets, and was only insignificantly worse on the fourth.
Additionally, ablating these features (Figure~\ref{fig: ablation_full}) showed that biases were always necessary to achieve this good performance, but adding correlation significantly hurt on two of the datasets and made an insignificant difference on the other two.
Thus, the combination of biases, effort, and censoring, which we found to be optimal in Section~\ref{sec: real_data} and used to generate data for our experiments in Section~\ref{sec: synthetic_data}, remained optimal after testing these additional changes.

\section{Additional Parameter Recovery Experiments}
\label{sec: parameter_recovery_appnedix}

\begin{figure}
    \centering
    \includegraphics[trim={0 0.4cm 0 0}]{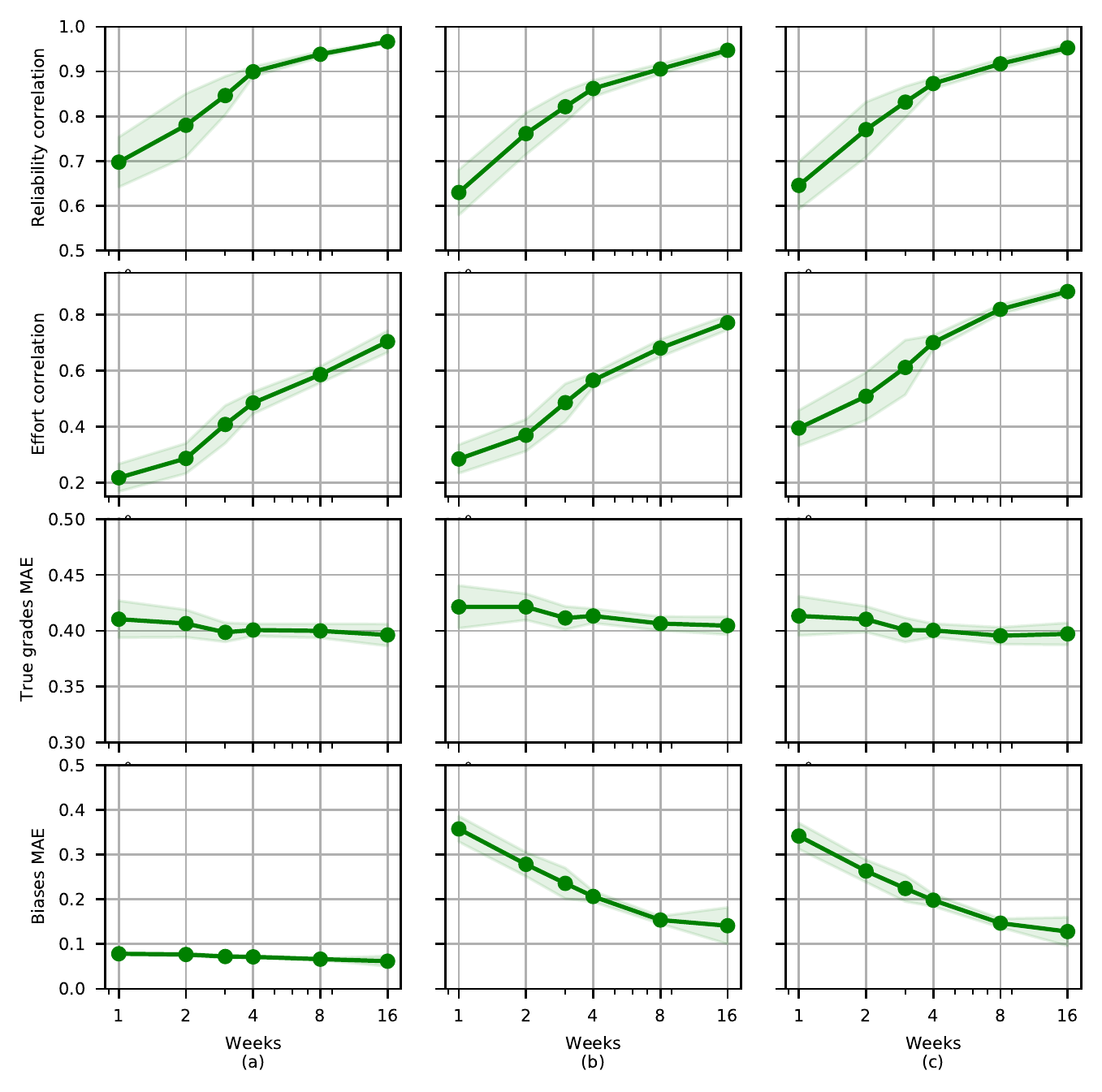}
    \caption{Effects of varying the number of weeks on parameter recovery.}
    \label{fig: effort_relaiblity_consistancy_rel_full}
\end{figure}

\begin{figure}
    \centering
    \includegraphics[trim={0 0.4cm 0 0}]{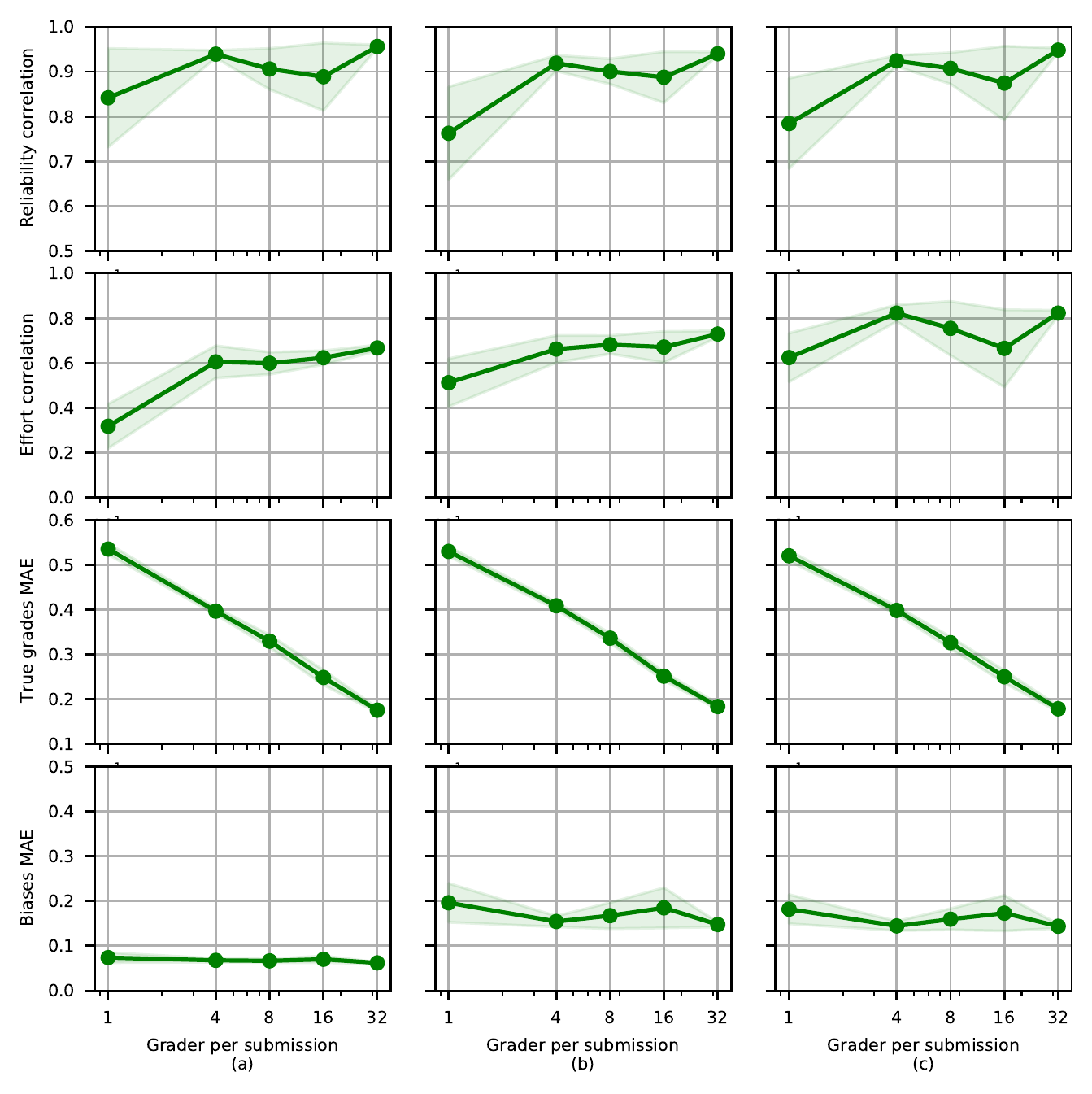}
    \caption{Effects of varying the number of grades per submission on parameter recovery.}
    \label{fig: true_grade_consistancy_rel_full}
\end{figure}

In this section, we include additional plots about our parameter recovery experiments on synthetic data.
In each figure, we include data from three different generating distributions, modeling (a) Fall18/19 (with $\sigma_b = 0.1$ and $\tau_\ell=1$), (b) Winter21 (with $\sigma_b = 1$ and $\tau_\ell=1$), and (c) Fall21 (with $\sigma_b = 1$ and $\tau_\ell=4$). 
Across all three of these generating distributions, we made the same conclusions: changing the amount of data available per student (Figure~\ref{fig: effort_relaiblity_consistancy_rel_full}) helped with reliability, effort, and bias estimates, while changing the amount of data available per submission (Figure~\ref{fig: true_grade_consistancy_rel_full}) helped with true grade estimates.

\section{Additional Misspecification Experiments}
\label{sec: misspec_appendix}
\begin{figure}
    \centering
    \includegraphics[trim={0 0.4cm 0 0}]{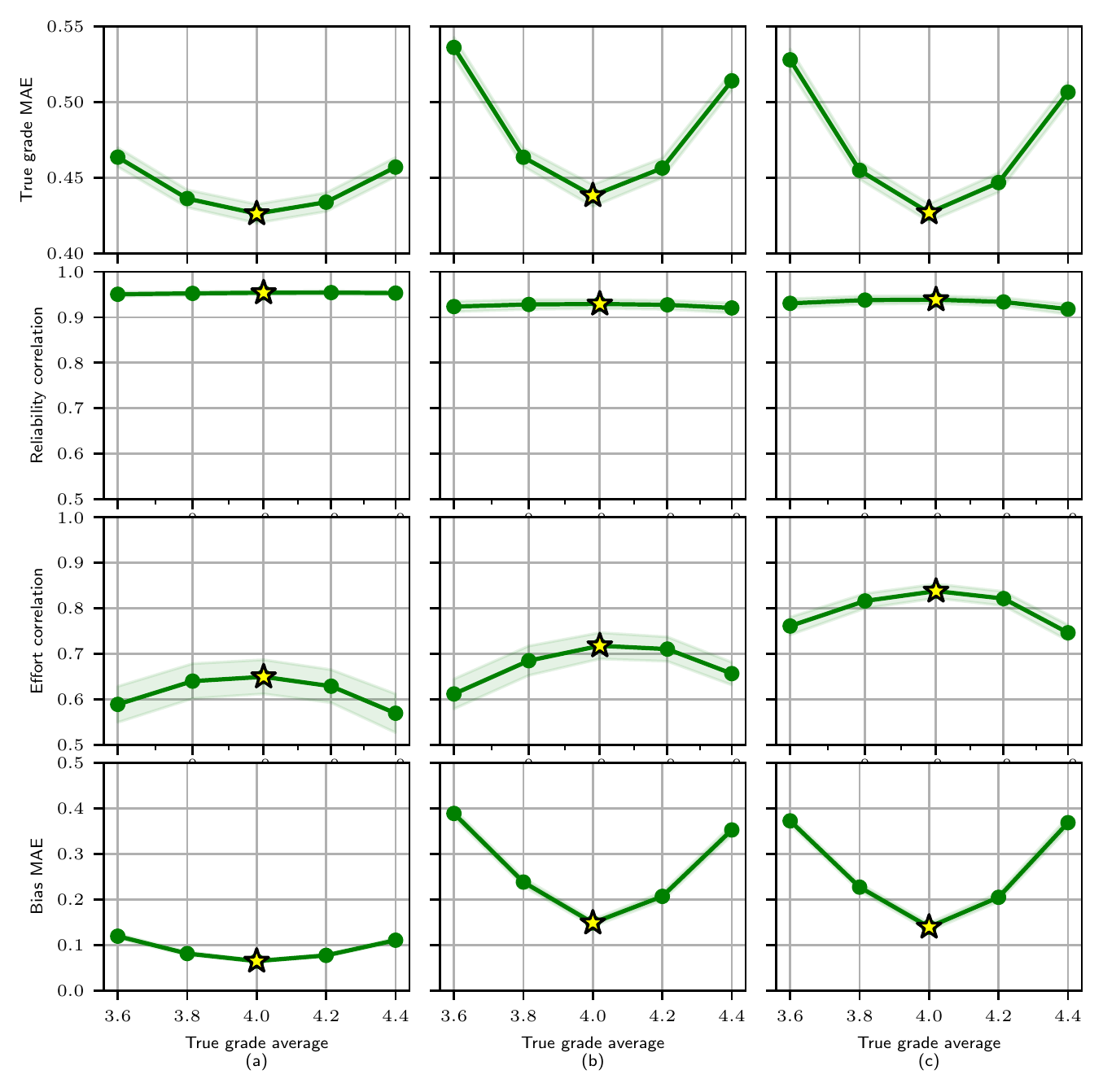} 
    \caption{Effect of misspecifying the true grade average $\mu_s$.}
    \label{fig: misspec_changing_mu_s}
\end{figure}

\begin{figure}
    \centering
    \includegraphics[trim={0 0.4cm 0 0}]{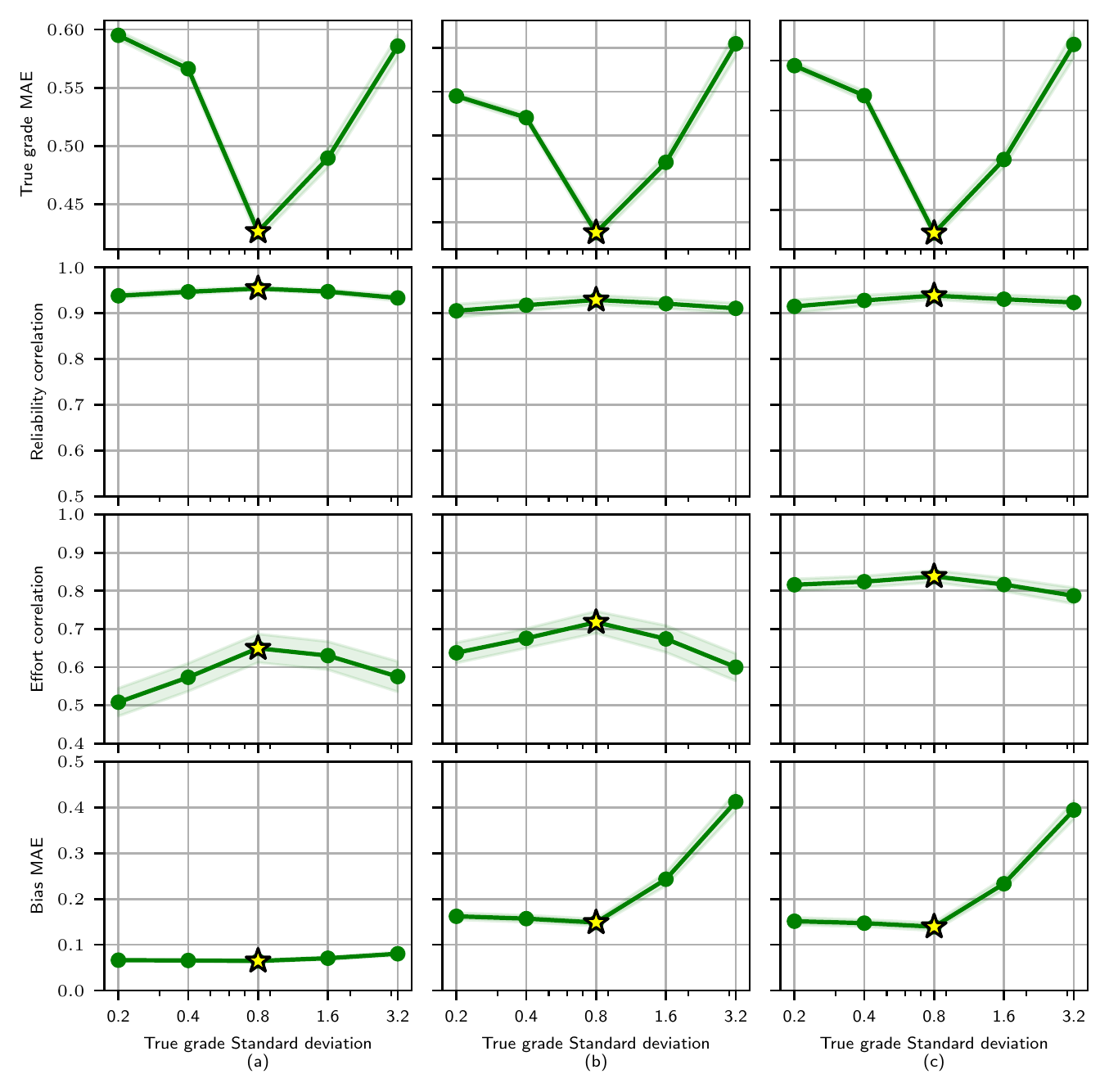} 
    \caption{Effect of misspecifying the true grade standard deviation $\sigma_s$.}
    \label{fig: misspec_changing_sigma_s}
\end{figure}

\begin{figure}
    \centering
    \includegraphics[trim={0 0.4cm 0 0}]{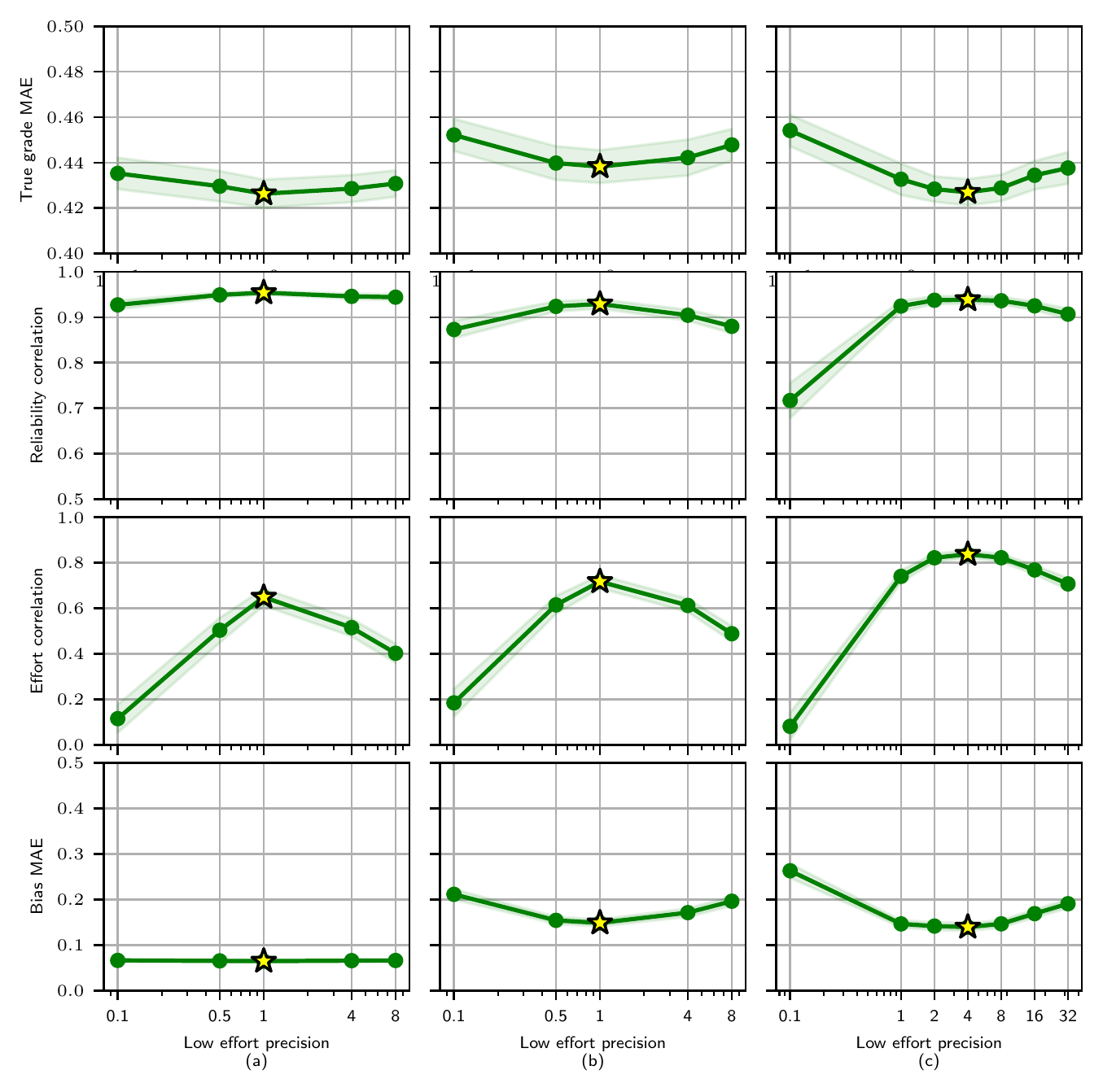} 
    \caption{Effect of misspecifying the true grade average $\tau_\ell$.}
    \label{fig: misspec_changing_tau_l}
\end{figure}

\begin{figure}
    \centering
    \includegraphics[trim={0 0.4cm 0 0}]{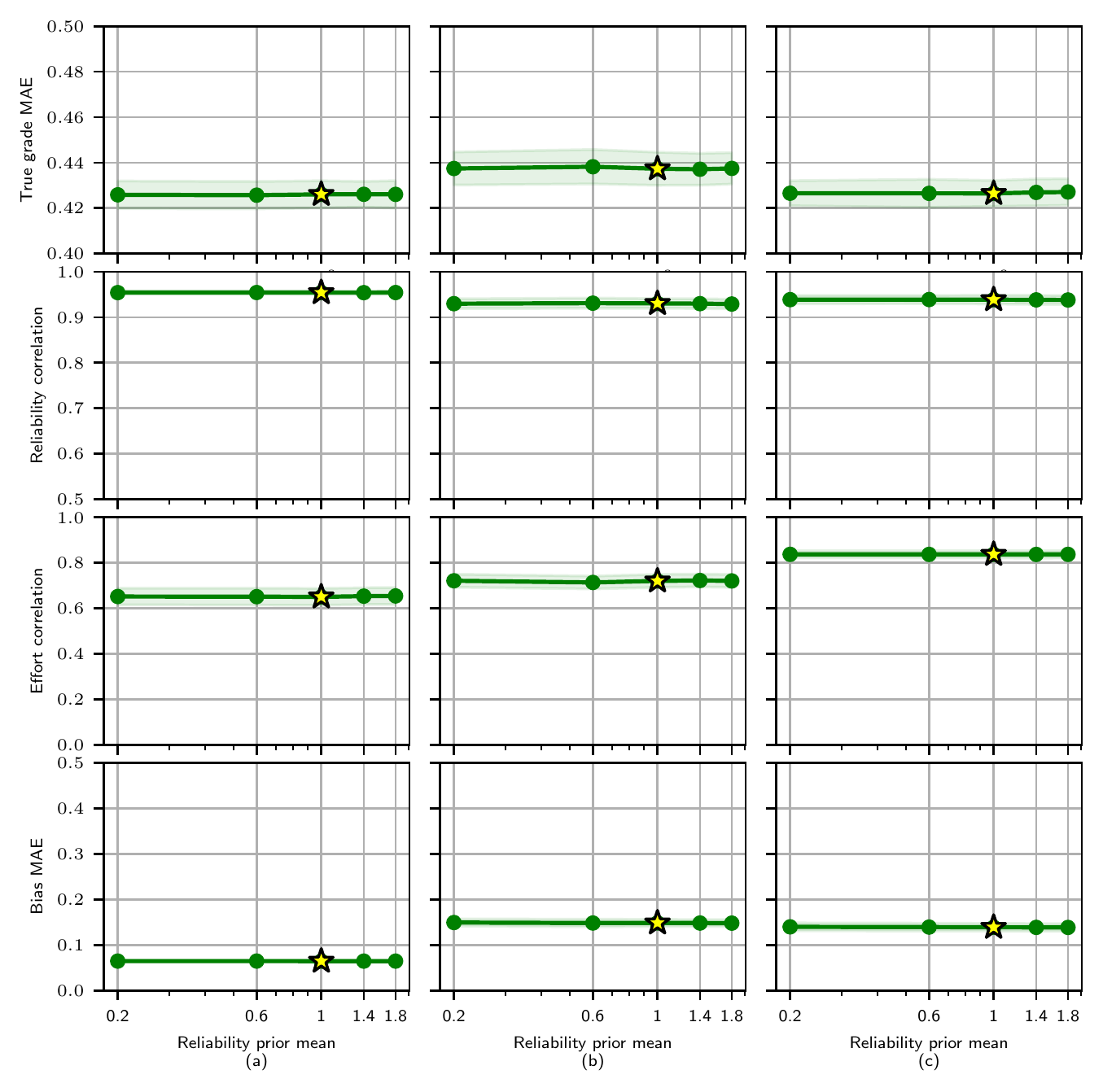}
    \caption{Effect of misspecifying the reliability prior. $\beta_\tau$ is held fixed; varying $\alpha_\tau$ changes both the prior's mean and the variance.}
    \label{fig: misspec_alpha_tau}
\end{figure}

\begin{figure}
    \centering
    \includegraphics[trim={0 0.4cm 0 0}]{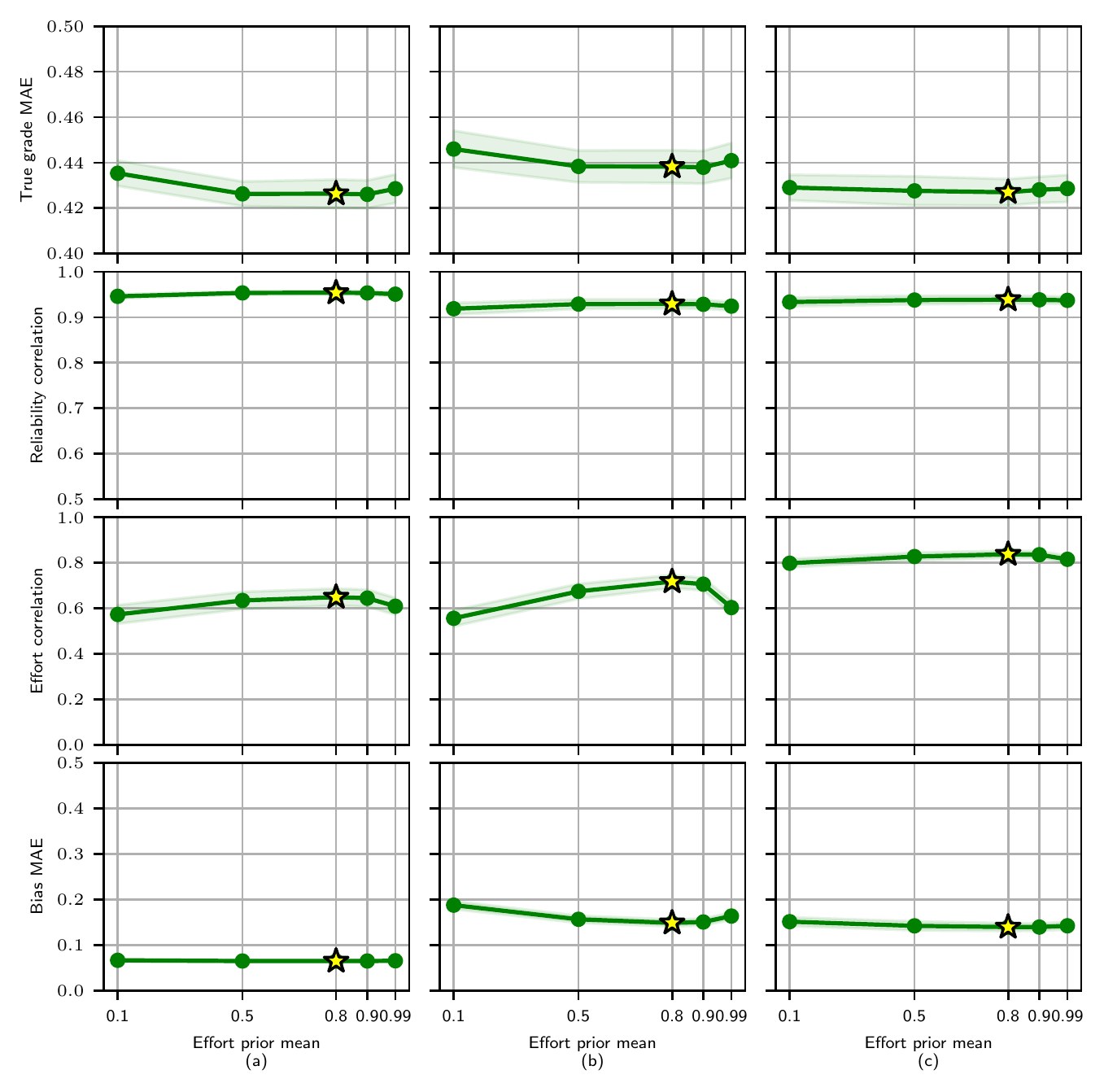} 
    \caption{Effect of misspecifying the effort prior mean $\alpha_e$ / $(\alpha_e + \beta_e)$, holding the prior's variance fixed.}
    \label{fig: misspec_alpha_e}
\end{figure}

\begin{figure}
    \centering
    \includegraphics[trim={0 0.4cm 0 0}]{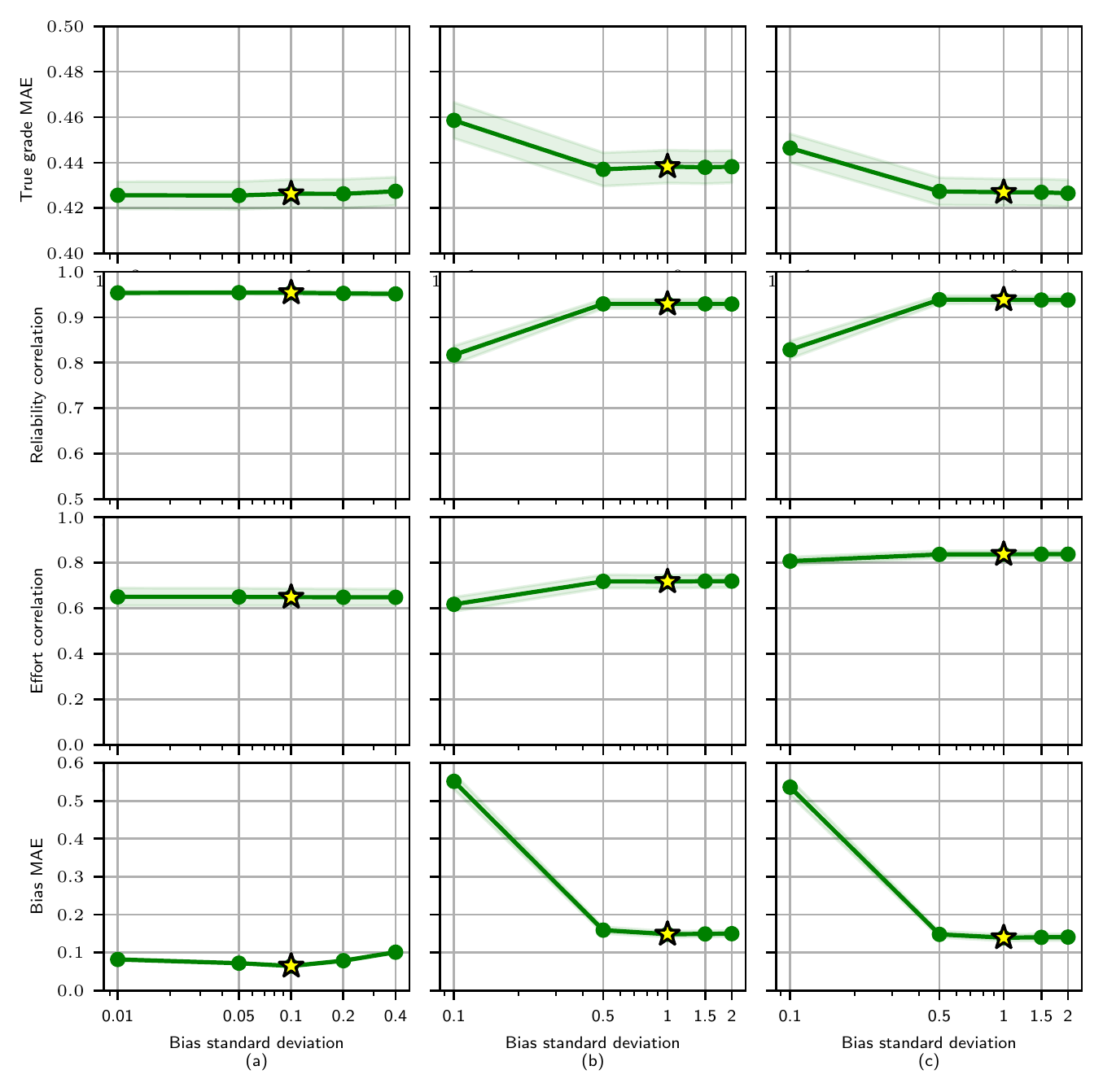} 
    \caption{Effect of misspecifying the bias standard deviation $\sigma_b$.}
    \label{fig: misspec_changing_sigma_b}
\end{figure}

In this section, we include additional plots about our misspecification experiments on synthetic data. 
Similar to Appendix~\ref{sec: parameter_recovery_appnedix}, we plot results using our distributions fit to (a) Fall18/19; (b) Winter21; and (c) Fall21 (with similar conclusions across datasets).
These plots show how true grade MAE, reliability Spearman correlation, effort Spearman correlation, and bias MAE vary as various hyperparameters are changed:
\begin{itemize}
    \item Figure~\ref{fig: misspec_changing_mu_s}: true grade average $\mu_s$;
    \item Figure~\ref{fig: misspec_changing_sigma_s}: true grade standard deviation $\sigma_s$;
    \item Figure~\ref{fig: misspec_changing_tau_l}: low-effort precision $\tau_\ell$;
    \item Figure~\ref{fig: misspec_alpha_tau}: reliability prior ($\alpha_\tau, \beta_\tau)$;
    \item Figure~\ref{fig: misspec_alpha_e}: effort prior ($\alpha_e, \beta_e)$;
    \item Figure~\ref{fig: misspec_changing_sigma_b}: bias standard deviation $\sigma_s$.
\end{itemize}

\end{document}